\documentclass[journal]{IEEEtran}

\usepackage{times}
\usepackage{epsfig}
\usepackage{graphicx}
\usepackage{amsmath}
\usepackage{amssymb}
\usepackage{array}

\usepackage{multirow}
\usepackage{subfigure}
\usepackage{diagbox}

\usepackage[pagebackref=true,breaklinks=true,letterpaper=true,colorlinks,bookmarks=false]{hyperref}

\newcommand{\etal}{\textit{et al}. }

\newcommand{\eg}{\textit{e}.\textit{g}. }
\newcommand{\wrt}{\textit{w.}\textit{r.}\textit{t.}}

\begin{document}

\title{Single Image Super Resolution - When Model Adaptation Matters}

\author{Yudong~Liang,
        Radu~Timofte,~\IEEEmembership{Member,~IEEE,}
        Jinjun~Wang,~\IEEEmembership{Senior Member,~IEEE,}
        Yihong~Gong,~\IEEEmembership{Senior Member ,~IEEE,}
        and~Nanning~Zheng,~\IEEEmembership{Fellow,~IEEE}% <-this % stops a space
\thanks{Yudong Liang, Jinjun Wang, Yihong~Gong and Nanning Zheng are with the Institute of Artificial Intelligence and Robotics, Xi'an Jiaotong University, China}% <-this % stops a space
\thanks{Radu Timofte is with Computer Vision Laboratory, Department of Information Technology and Electrical Engineering, ETH Z\"urich, Switzerland.}% <-this % stops a space
\thanks{}}

\markboth{Journal of \LaTeX\ Class Files,~Vol.~14, No.~8, August~2015}%
{Shell \MakeLowercase{\textit{et al.}}: Bare Demo of IEEEtran.cls for IEEE Journals}

\maketitle

\begin{abstract}
In the recent years impressive advances were made for single image super-resolution. Deep learning is behind a big part of this success. Deep(er) architecture design and external priors modeling are the key ingredients. The internal contents of the low resolution input image is neglected with deep modeling despite the earlier works showing the power of using such internal priors.
In this paper we propose a novel deep convolutional neural network carefully designed for robustness and efficiency at both learning and testing. Moreover, we propose a couple of model adaptation strategies to the internal contents of the low resolution input image and analyze their strong points and weaknesses. By trading runtime and using internal priors we achieve 0.1 up to 0.3dB PSNR improvements over best reported results on standard datasets. Our adaptation especially favors images with repetitive structures or under large resolutions. Moreover, it can be combined with other simple techniques, such as back-projection or enhanced prediction, for further improvements.
\end{abstract}


% \fi
\IEEEpeerreviewmaketitle

\section{Introduction}

\IEEEPARstart{S}{ingle} image super-resolution (SR) is a fundamental yet challenging vision problem of high practical and theoretical value. It is a particularly interesting problem as 4K images, videos and displays are of huge demands nowadays and most digitally recorded media used lower resolutions. Single image super-resolution aims at reconstructing a high resolution (HR) image for a given low resolution (LR) input. It is an ill-posed problem as usually to a low-resolution image patch there is a large number of corresponding high-resolution image patches. Numerous attempts have been made to solve this under-constrained or ill-posed problem. The techniques can be roughly divided into three categories: the interpolation methods~\cite{keys1981cubic}, the reconstruction methods~\cite{irani1993motion,aly2005image}, and the example based methods~\cite{freeman2000learning,yang2008image,timofte2013anchored}. These methods exploit priors ranging from simple `smoothness' priors applied in interpolation to more sophisticated statistical priors learned from large collections of natural images or from internal examples extracted from the low resolution image. Local structures, patterns, or textures tend to reappear many times across natural images~\cite{freeman2000learning,yang2008image,timofte2013anchored}, and even redundantly recur within the same image of the same scale or across different scales of the image~\cite{glasner2009super,freedman2011image,zontak2011internal}.
\begin{figure}[th!]
     \centering
     \footnotesize
     \includegraphics[width=0.490\textwidth]{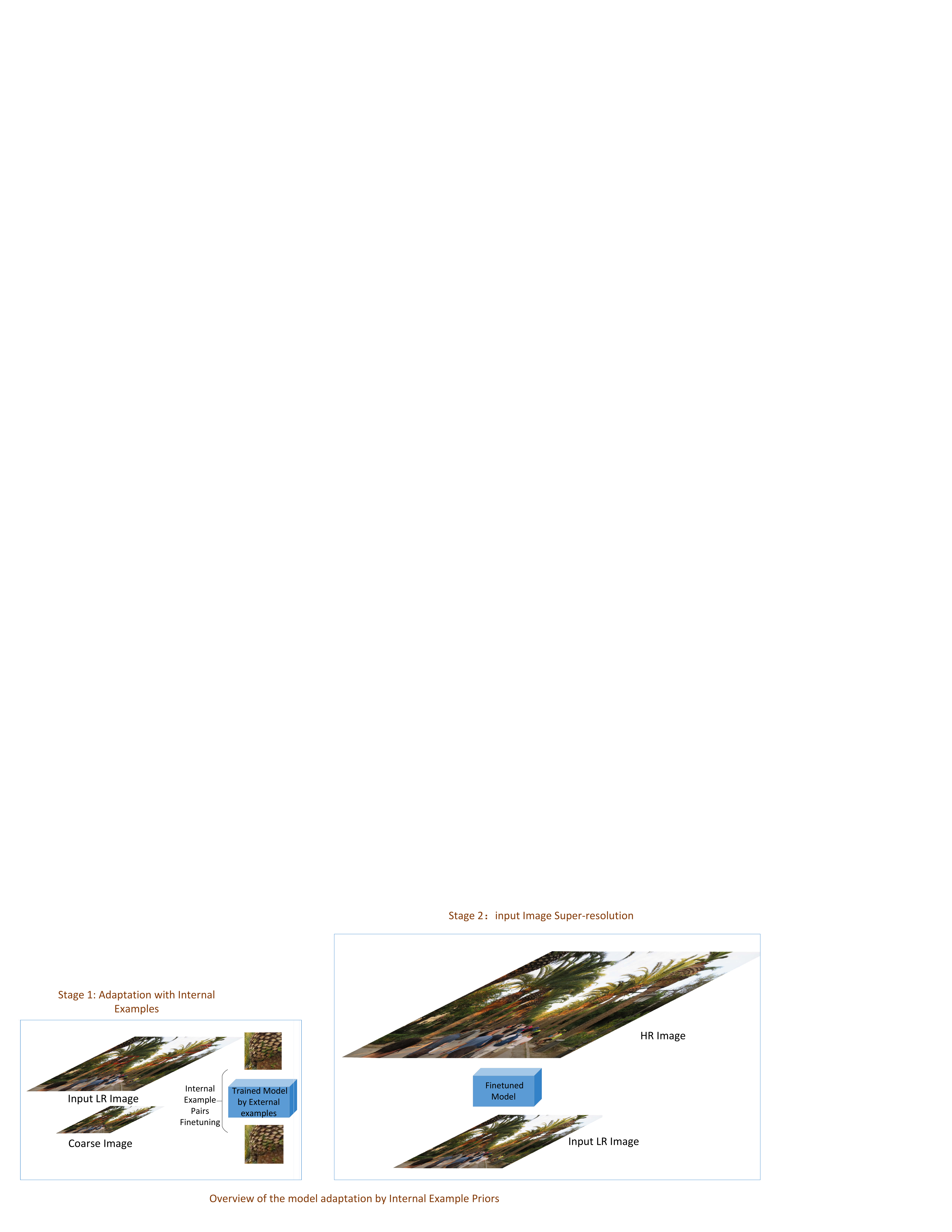}
     \caption{Overview of our model adaptation strategy by internal example priors. The finetuned model has been trained using external examples.}
  \label{fig:overview}
  \end{figure}
The recurrence priors between LR and HR image examples are modeled by learning a mapping from internal examples~\cite{freeman2000learning,glasner2009super}, external examples~\cite{chang2004super, timofte2013anchored,dong2016image} or from combined sources~\cite{yang2013fast,timofte2016seven}.

The example based methods are the main research direction of the recent years. They refer to the priors from examples to restore missing details (high frequencies) in addition to the LR input information (low frequencies). These priors regulate the ill-posed super-resolution problem and help to obtain sharper and more visual pleasing results.
In order to cover the high diversity of contents present in the natural images the external priors are usually extracted from large collections of LR/HR image example pairs. As demonstrated by many works~\cite{kim2016accurate,timofte2016seven,dong2014learning}, the amount of external train examples has a strong influence on the performance of the trained methods. On the other side, although the number of internal examples which come from the same input LR image or the scaled version is limited, the predictive power of the internal priors~\cite{glasner2009super,freedman2011image,zontak2011internal} is often stronger than that of the external priors.
The internal priors are specific to the LR image and match the contextual and semantic information and the contents, while the external priors often embed generic and sometimes irrelevant information for a specific LR image.
Patch redundancy or self-similarity probability of (small) image patches within the same scale or across a pyramid of scales of a natural image generally increases with the size of the image.
With the development of the camera sensors the sizes of the photos continuously increase and indicate potential for exploiting internal priors as more internal examples can be used for an image specific adaptation SR process.

Internal priors are also widely applied in other image restoration applications, such as image denoising~\cite{burger2013learning,dong2013nonlocally}. Yang~\etal~\cite{yang2013fast} proposed a regression model for image superresolution by leveraging external examples and internal self-examples.
 The power of combination of internal priors and external priors calls for a effective way to further improve the superresolution restoration.

Among example based methods, the learning based methods such as sparse coding~\cite{yang2008image}, neighbor embedding~\cite{chang2004super,timofte2013anchored,timofte2014a+} and recent deep learning based methods~\cite{dong2014learning,kim2016accurate} have largely promoted the performance of the SR process. Dong~\etal~\cite{dong2014learning,dong2016image} proposed for SR the first successful deep convolutional neural network (CNN) termed SRCNN of three convolutional layers. SRCNN indicates the importance of feature representation which enjoy the merits of CNNs.
However, SRCNN have failed to improve the performance by increasing the depth due to the difficulty of training. Inspired by the great success of deeper models in other computer vision tasks like image classification~\cite{he2015deep,simonyan2014very,he2016identity}, Kim~\etal~\cite{kim2016accurate} proposed a very deep CNN (VDSR) with depth up to 20 to predict the residual image. It applies a simple plain stacking network to predict the residual between the HR and LR images, which largely boosted the convergence speed and performance. It demonstrated again that the representational ability is very important to the single image SR problem.

Although VDSR has achieved impressive results, the plain structure of VDSR which simply stacks layers suffers gradient exploding/vanishing problem as network goes deeper. The convergence of deeper architectures becomes more difficult. The success of residual networks~\cite{he2016identity,he2015deep} indicates the importance of skip connection in the information propagation. Another intuitive observation is that the simple stacking of too many ReLU layers has lost too much information. Dong~\etal~\cite{dong2016accelerating} has reported the use of PReLU has alleviated the `dead feature' problem caused by zero gradients of ReLU layers. Alternatively, we propose a novel projection skip connection to preserve the negative information and to alleviate the gradient exploding/vanishing problem. With the idea to preserve negative information, a modification of reconstruction parts that discards ReLU layers has been proposed to cooperate with information propagation of projection skip connection. To avoid the overfitting and computational problem, we propose a Deep Projection convolution neural Network (DPN) with progressive pyramidal architecture which despite going deeper (40 layers vs. 20 layers) does not surpass the number of parameters from VDSR.

With DPN, our new design of deep architecture, we obtain state of the art performance with a learned model on external examples. The difficulty for training has been largely alleviated by our architecture. For a certain amount of data, the discriminative ability of the data to the problem is highly relevant to performance of the model. We hope to arouse the interests of communities for model adaptation with internal examples. To do model adaptation to a given image we propose a simple but effective method --  finetuning of the external and offline learned model with internal examples as depicted in Fig.~\ref{fig:overview}.
The performance is further boosted with such an adaptation to each test image. Note that in all our experiments the adapted models perform comparable or better than the externally trained one.
We validate the generality of our model adaptation strategy also when starting from VDSR.

In this paper, we make the following main contributions:
\begin{enumerate}
\item we present a novel deep model -- DPN with performance, robustness and easiness to train merits \wrt state-of-the-art VDSR model;
\item we propose and study a couple of model adaptation techniques as effective ways to significantly improve the performance of single image super-resolution methods;
\item we contribute an accurate model selection strategy based on observed image cross-scale content similarity and model performance correlation.
\end{enumerate}
The remainder of the paper is structured as follows. Section~\ref{sec:proposedmethod} introduces our deep architecture and the proposed methods. Section~\ref{sec:experiments} describes the experiments and the achieved results. Section ~\ref{sec:discussion} tries to share some insights and take-home messages. Section~\ref{sec:conclusion} concludes the paper.

\begin{figure*}[th]
     \centering
     \footnotesize
     \includegraphics[width=0.950\textwidth]{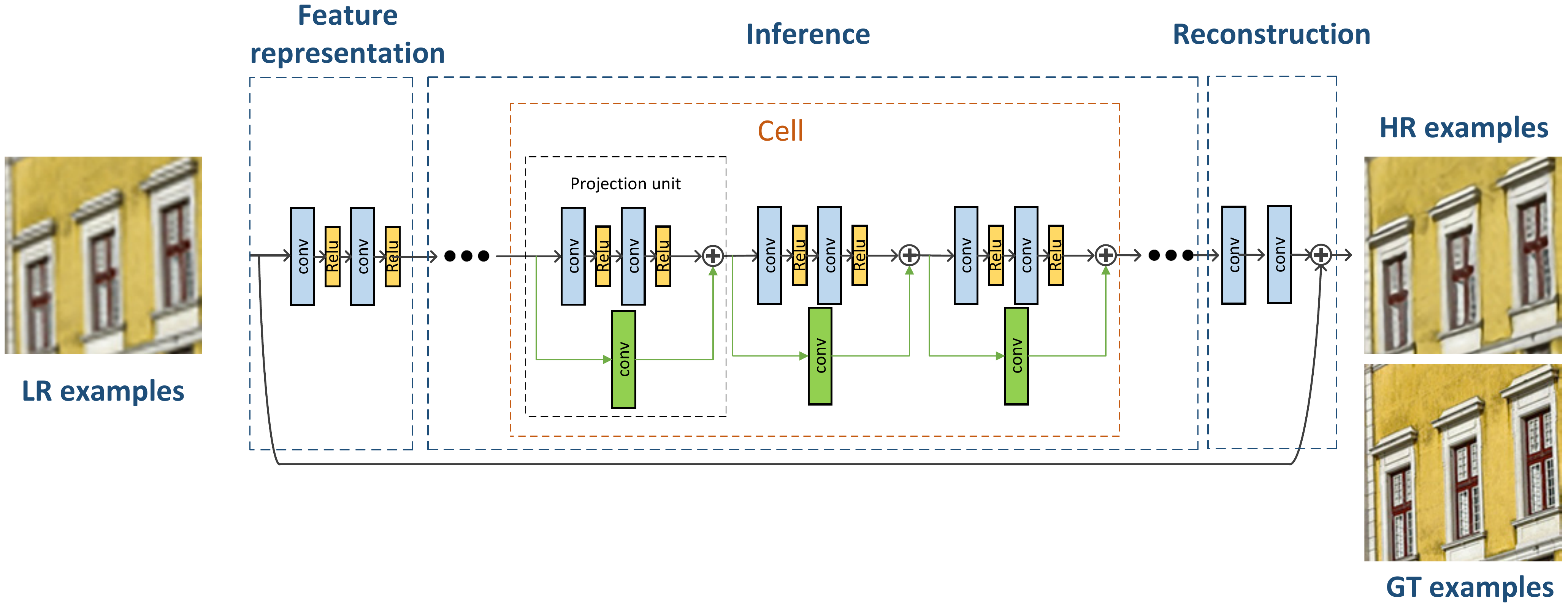}
     \caption{The architecture of our deep projection convolutional neural network (DPN).}
  \label{fig:DPN}
  \end{figure*}

\section{Proposed method}
\label{sec:proposedmethod}

First, we introduce the architecture of our Deep Projection convolutional neural Network (DPN) and discuss the motivations behind its design. Then, we propose a couple of model adaptation strategies to the internal contents of the LR input image, including a model selection strategy from adaptation, and analyze their strong points and weaknesses.

\subsection{Motivation of the proposed deep architecture}

Single image super-resolution also enjoys the performance benefits of deeper architectures. However, with plain stacking of convolutional layers and ReLU layers come also training difficulties~\cite{he2015deep}. Gradient exploding/vanishing problem is the main issue. Stacking the convolutional layer with ReLU layer outputs non-negative values which may impact the representational ability in SR. Moreover, overfitting may exist for certain low amount of train data. We tackle these problems by proposing our novel Deep Projection convolutional neural Network (DPN) depicted in Fig.~\ref{fig:DPN}.
DPN has three parts: feature extraction/representation, inference, and reconstruction. The feature extraction part are plain stacking convolutional and ReLU layers as usual, the inference part consists of projection units and the reconstruction only applies convolutional layers to restore HR images.
The projection design (see Fig.\ref{fig:Comp}a)) is meant to reduce the training difficulties and to preserve information as complete as possible at the same time, which makes the training more efficient and more robust.

Inspired by the deep residual CNN~\cite{he2015deep} and identity mapping deep CNN~\cite{he2016identity}, gradient vanishing problem can be largely alleviated by skip connections or shortcuts. The calculation of residual unit can be as follows

\begin{equation}
\label{eq:identity}
x_{k+1}=x_k+F(x_k,W_k)
\end{equation}
where $x_k$ is the input of the unit and $F$ is a function of $x_k$ with parameters $W_k$.

The shortcuts decrease the difficulty of training but identity shortcuts are not sufficient to make up for the information loss due to non-negative outputs. Intuitively, the outputs need a refinement such
             \begin{equation}
             \label{eq:proj}
                 x_{k+1}=G(x_k)+F(x_k,W_k)
             \end{equation}
 where $G$ can be a function with an output in $[-\infty,\infty]$. The G behaves like a shortcut which is expected to be simple. A natural choice is a convolutional operation as in Fig.\ref{fig:Comp}a). The non-negative output $F(x_k,W_k)$ brings non-linearity and projection $G(x_k)$ extends the descriptive ability. Thus, the inference part preserves information and alleviates the training difficulties. In combination with inference part, stacking convolutional layers only is applied in the reconstruction parts, otherwise with ReLU layers losing too much information at the end may impact the performance. Our designing philosophy coincides with~\cite{dong2016accelerating} which uses PReLU for increasing the representation power. Our reconstruction part makes our net more stable, more robust and training more efficient as showed by the experiments.

The overfitting problem arises when the number of learned model parameters is too large for a given amount of train data. At the same time, for a good descriptive ability the network needs to be deep with a sufficient number of filters. In our DPN we increase the number of filters in a progressive pyramidal fashion and go deeper but without more parameters than VDSR.

The motivations to design the introduced CNN architectures include: 1) it adopts projection skip connections to allow for deeper structure; 2) the projection unit introduces extra convolution to increase the model capacity and preserves negative information; 3) integrating reconstruction part reduces the information loss; and 4) the design of the net complexity in a progressive pyramidal fashion controls the parameter numbers to be no more than VDSR.
Experimentally, our implementation outperforms our reimplementation of VDSR for more than 0.1dB, which validates the strength of our proposed model structure.
%-------------------------------------------------------------------------
\subsection{Architecture}

\begin{figure*}[th]
  \centering
  \footnotesize
  \begin{tabular}{cccc}
   \subfigure{\includegraphics[width=0.220\textwidth]{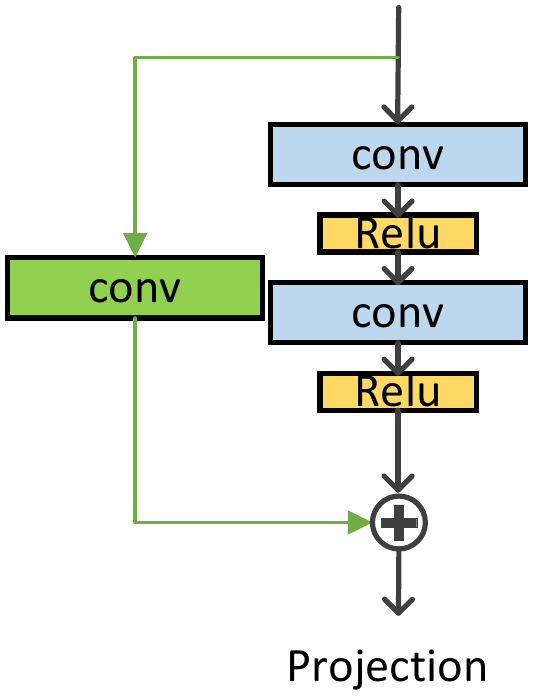}}&
  \subfigure{\includegraphics[width=0.220\textwidth]{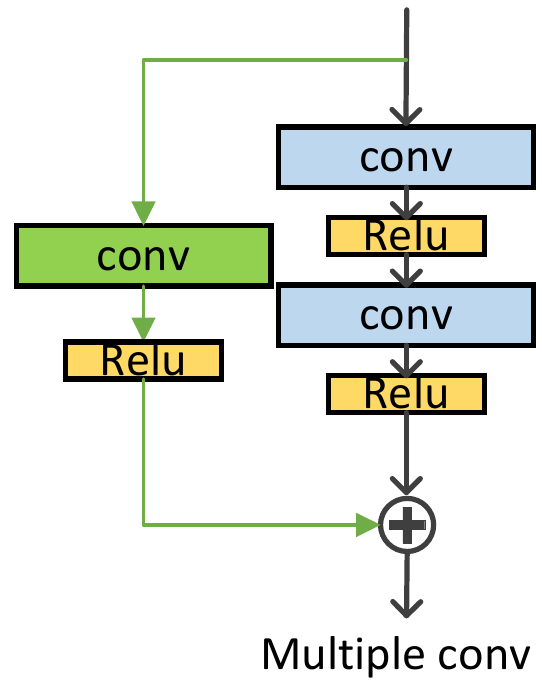}}&
  \subfigure{\includegraphics[width=0.220\textwidth]{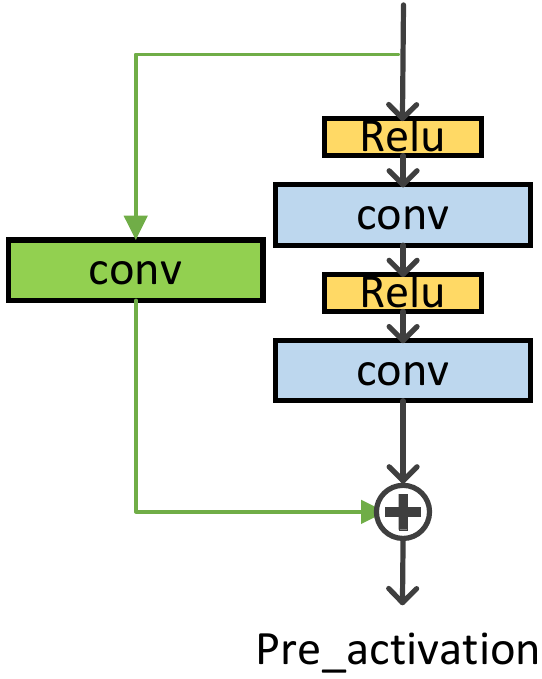}} &
    \subfigure{\includegraphics[width=0.190\textwidth]{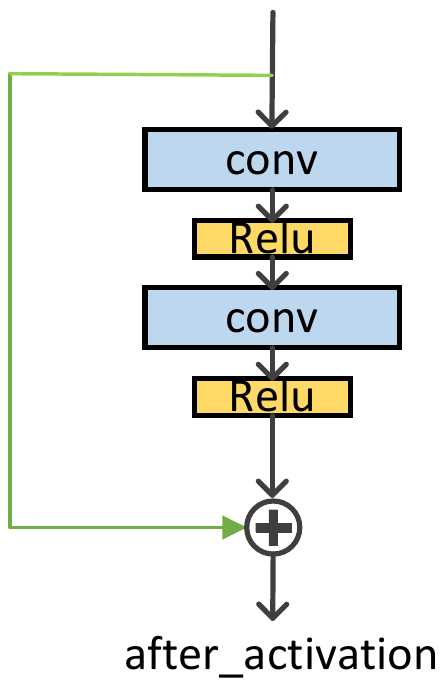}} \\
    (a)&(b)&(c)&(d)\\
  \end{tabular}
  \caption{Different architecture cells of CNN}
  \label{fig:Comp}
\end{figure*}

Our Deep Projection convolutional neural Network (DPN) is composed of several \textit{Cells} which have certain number of projection units as shown in Fig.~\ref{fig:DPN}. Each unit consists of a summation between the output of 2 stacked conv+ReLU and the output of a conv layer as in Fig.~\ref{fig:Comp}. All the convolutional layers have a receptive field of $3\times3$. For succinctness in our implementation:

\begin{itemize}
  \item The filter and channel numbers of the each projection unit in the cell $i$ are the same, denoted as $N_i$.
  \item The numbers of the projection units in each cell are the same, $k$.
  \item Each cell is described as a collection of projection units, $\{N_i\}^k$
  \item  The inference part is a collection of cells as \{$\{N_1\}^k$, $\{N_2\}^k$, $\cdots$\}.
\end{itemize}

DPN has 40 convolutional layers altogether. The inference part is composed of 6 cells with increasing number of projection units: $(16^3, 16^3, 32^3, 32^3, 64^3, 64^3)$. The extraction and reconstruction part has 16 filters. Each projection unit has a depth 2.
Due to the progressive design of pyramidal shape our DPN is much deeper (40 layers vs. 20 layers) but has no more parameters than VDSR.

 \subsection{Finetuning with internal priors}
Internal priors are a powerful source of information but barely applied for deep learning architectures mainly due to the lack of internal examples and expensive training. We propose the adaptation to the internal priors of an input LR image as a finetuning to internal examples of the externally trained deep model. This strategy is applicable to other deep models such as VDSR.

\textbf{Internal examples extraction}
The natural images exhibit a local recurrence of structures and textures within same and across different (coarser) image scales. This motivates our internal examples extraction. Each test LR image is downscaled to form a pyramid of scaled images from which internal example LR-HR pairs are extracted.

\textbf{Finetuning}
Our DPN model trained with external examples is first finetuned to the internally extracted examples such that to adapt to the internal priors of each test LR image. The super-resolved HR output is obtained using the finetuned DPN model. Different LR input images will have different adapted models. It is a time expensive but performance superior approach. In our experiments, if \textit{sufficient internal examples} or \textit{strong self-similarities} exist, the model will be largely improved even within one training epoch.

  \begin{figure}[th]
  \centering
  \footnotesize
  \begin{tabular}{cc}
   \subfigure{\includegraphics[width=0.220\textwidth]{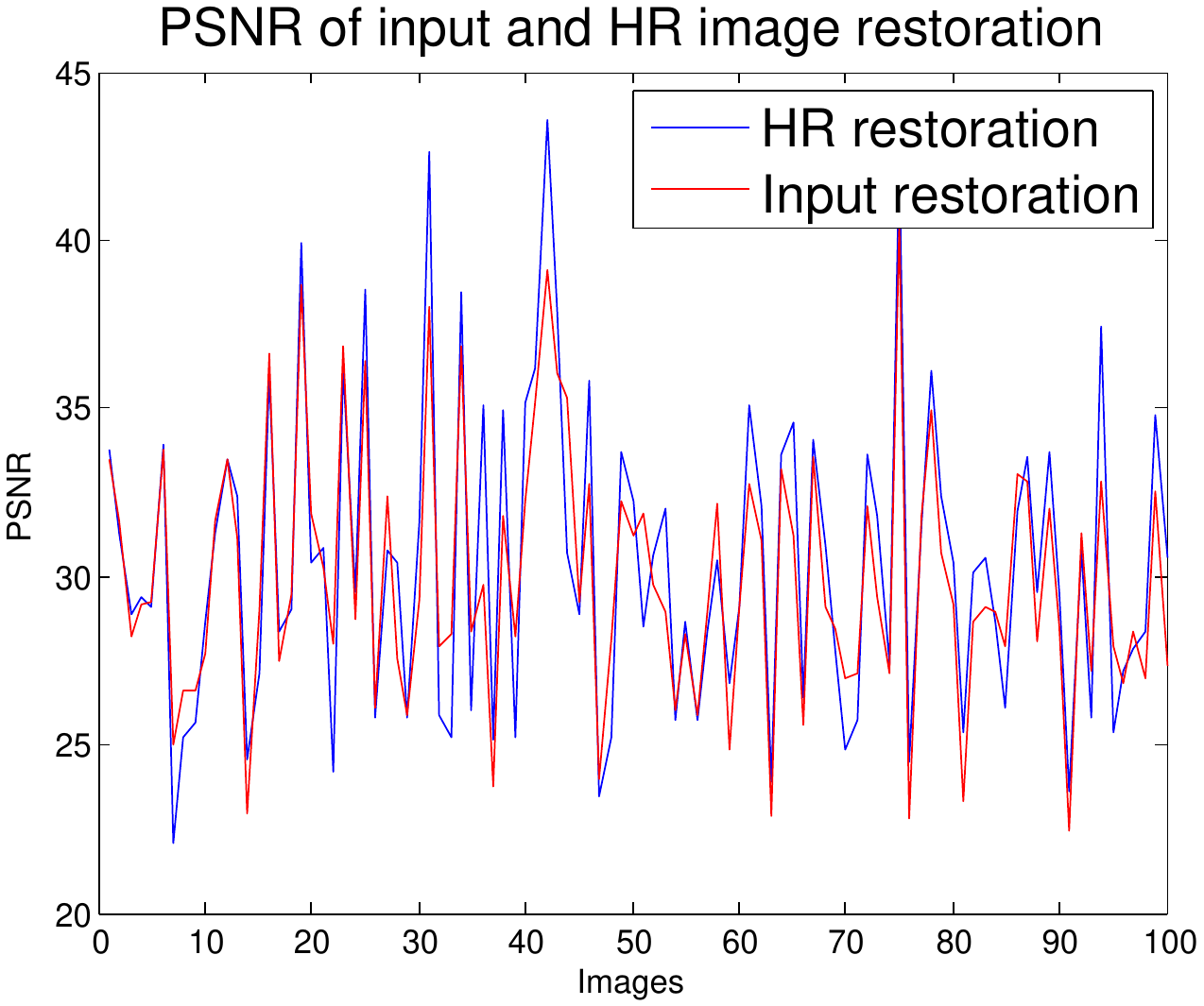}}&
  \subfigure{\includegraphics[width=0.220\textwidth]{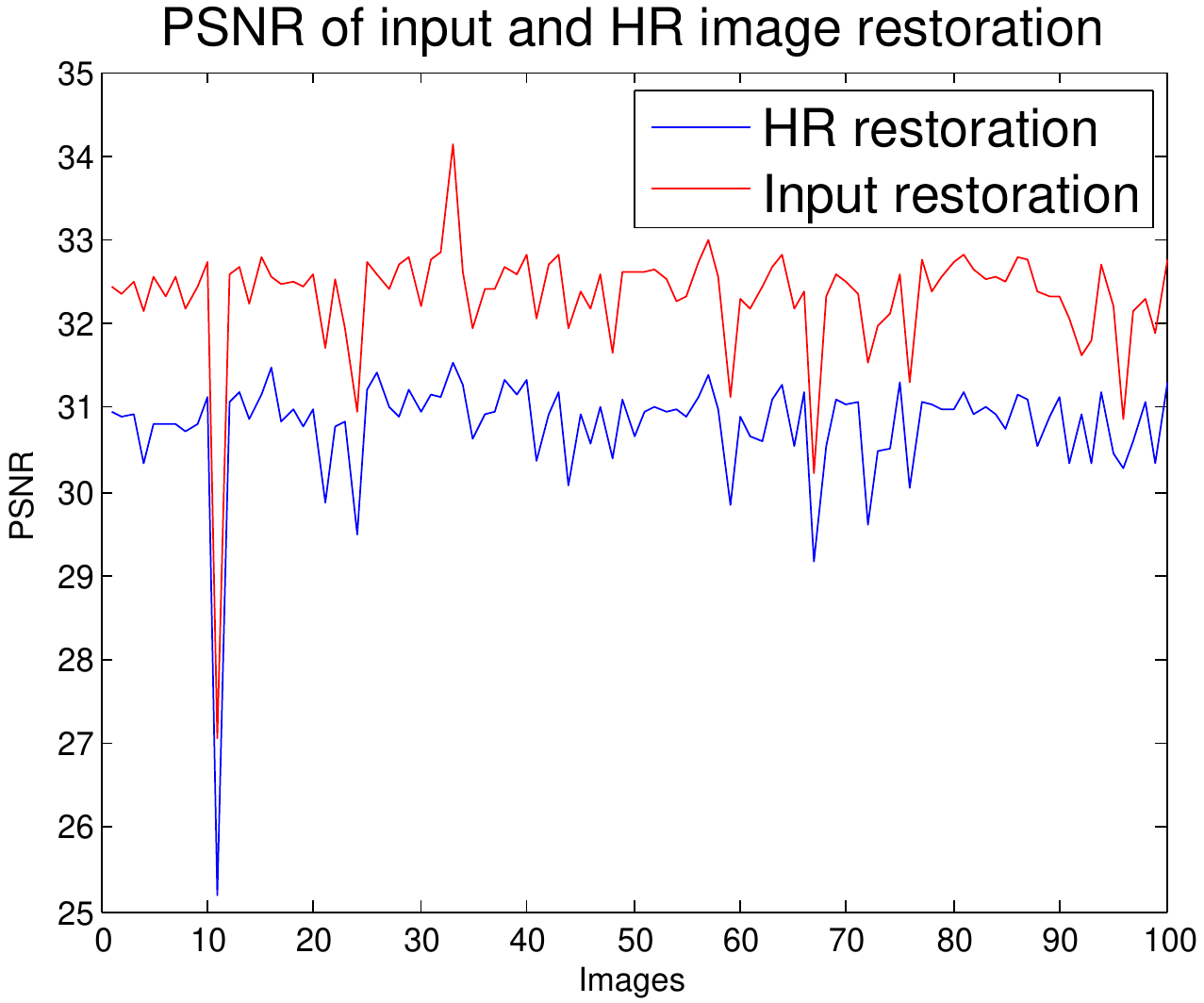}} \\
    (a)&(b)\\
  \end{tabular}
  \caption{PSNR (dB) curves of the input image restoration and HR image super-resolution on Urban100. The left corresponding to 100 randomly selected results from the 10000 restoration pairs, while the right corresponding to 100 restorations of 1 image (randomly selected).}
  \label{fig:lr2hr}
\end{figure}

\subsection{Adaptation as a model selection}
\label{sec:adap_select}
When facing with a test image, one can predict the HR image super-resolution performance of a model by evaluating the performance of input image restoration by this model. For Urban100 dataset, 100 models are finetuned for 100 LR images, respectively. For each LR image and the further downscaled image of it, we apply all these models to obtain the HR image and the restored LR input, respectively. In Fig.~\ref{fig:lr2hr} we randomly selected 100 image and model pairs and show the achieved restoration performance in terms of Peak Signal to Noise Ratio (PSNR) dB. We note that the performances in each of the 100 pairs are highly correlated between starting from the LR input and starting from a further downscaled LR input, the `HR restoration' and `Input restoration' labels in Fig.~\ref{fig:lr2hr}. Thus, one can often predict the performance of a model on an image by checking the performance of the model on the downscaled image. Our adaptation as a model selection strategy is as follows:
     \begin{itemize}
       \item  Offline train and store finetuned DPN (or other) models for hundreds of (diverse) images.
       \item  Online apply the stored DPN (or other) models to a downscaled image of the input and check which of the models perform the best.
       \item  Super-resolve the input LR image with the best selected model(s).
     \end{itemize}

This adaptation as a model selection strategy is particularly useful when the test image is small (thus few internal examples) or when the finetuning/training is too expensive / infeasible. Often running a (deep) model is orders of magnitude faster than adapting it by finetuning with internal examples.

\section{Experiments}
\label{sec:experiments}

In this section, we conduct a series of experiments to compare the performance of our proposed DPN method against the state-of-the-art single image SR methods, especially to demonstrate the power of internal example adaptation. All the experiments use the deep learning toolbox matconvnet~\cite{arXiv:1412.4564} and a single NVIDIA K40/K80 GPU card.~\footnote{We will publicly release both code and models upon paper acceptance.}

\subsection{Experimental settings}
\textbf{Train dataset.} We use 291 train images as in~\cite{kim2016accurate,schulter2015fast}, which include 91 images from Yang~\etal~\cite{yang2008image} with the addition of 200 images from Berkeley Segmentation Dataset (BSD)~\cite{martin2001database}.

\textbf{Test datasets} Five datasets are investigated: Set5 and Set14 from~\cite{timofte2013anchored}, Urban100 from~\cite{huang2015single} and BSD100 from~\cite{timofte2014a+,yang2014singleBenchmark} and L20 from~\cite{timofte2016seven}. Specifically, Urban100 is famous for its self-similarities, L20 has very large images, between 3m pixels to up to 29m pixels, while the other datasets have images below 0.5m pixels.

As in~\cite{timofte2013anchored,dong2014learning,kim2016accurate,schulter2015fast} we only super-resolve the luminance channel in YCrCb colorspace in our experiments and use the same downsizing operator (`imresize' from Matlab with `bicubic' interpolation) for obtaining the LR images from the ground truth HR images. All the training samples are randomly rotated or flipped as in~\cite{timofte2016seven,kim2016accurate} with a size of $41\times41$. DPN training uses batches of size 64, momentum and weight decay are set to 0.9 and 0.0001, respectively. We train a single model for multiple scales with the same strategy as VDSR, which is mixing samples from all scales ($\times 2$,$\times 3$, and $\times 4$) randomly. The gradient clipping is applied as in~\cite{kim2016accurate}. Learning rate is initially set to 0.1 and then decreases by a factor of 10 every 30 epochs in logarithm manner which allow a smooth decreasing.

For internal finetuning, larger magnification factors can be achieved by upscaling small factors multiple times in a cascaded fashion~\cite{timofte2016seven,wang2015deep}. For internal example adaptations we report mainly for $\times 2$ magnification factor.
Finetuning learning rate is fixed as $1e-4$ during model adaptation.

\subsection{ DPN with external examples }
The performance of our DPN trained on external examples is quantitatively evaluated in terms of PSNR (dB) and SSIM measures on 4 datasets and for 3 upscaling factors. In Table~\ref{table:Proj_basic} we report the results of our DPN in comparison with A+~\cite{timofte2014a+} (a state-of-the-art neighbor embedding method), RFL~\cite{schulter2015fast} (a random forest approach using local regressors as in A+), SelfEx~\cite{huang2015single} (a self-exemplars method using internal priors), SRCNN~\cite{dong2016image} (revisited deep learning method), and VDSR~\cite{kim2016accurate}, the very deep and current state-of-the-art CNN method. For all the compared methods we used the reported results or results achieved by original codes and models as provided by the authors. Bicubic interpolation results are provided for reference. In most of the settings our DPN achieves the best performance, especially for the $\times 3$ and $\times 4$ upscaling factors, while in the other cases, DPN is slightly (0.01dB) below VDSR.

Note that with our VDSR reimplementation and retraining we were not able to achieve the reported VDSR performance (more than 0.1dB gap in Set14). We hope that with the missing details of VDSR our model can be further improved.

 %//////////////////model performance before adaptation/////////////////////////////
   \begin{table*}[bht!]%[htb!]
        \centering
         \caption{Methods comparison on 4 datasets for 3 upscaling factors in terms of average PSNR and SSIM measures.}
         \label{table:Proj_basic}
         %\resizebox{\linewidth}{!}
         {
         \begin{tabular}{c c|c c c c c c c c}
         \multirow{2}{*}{Dataset} & \multirow{2}{*}{Scale} & Bicubic & A+\cite{timofte2014a+} & RFL\cite{schulter2015fast} & SelfEx\cite{huang2015single} & SRCNN\cite{dong2016image} & VDSR\cite{kim2016accurate} &\textbf{DPN (ours)}\\
         && PSNR/SSIM & PSNR/SSIM & PSNR/SSIM & PSNR/SSIM & PSNR/SSIM & PSNR/SSIM &  PSNR/SSIM\\
         \hline
         \hline
         \multirow{3}{*}{Set5} & $\times$2 & 33.66/0.9299 & 36.54/0.9544 & 36.54/0.9537 & 36.49/0.9537 & 36.66/0.9542 & \textbf{37.53}/\textbf{0.9587} &    37.52 /0.9586  \\
         & $\times$3 & 30.39/0.8682 & 32.58/0.9088 & 32.43/0.9057 & 32.58/0.9093 & 32.75/0.9090 & 33.66/0.9213 &   \textbf{33.71}/\textbf{0.9222} \\
         & $\times$4 & 28.42/0.8104 & 30.28/0.8603 & 30.14/0.8548 & 30.31/0.8619 & 30.48/0.8628 & 31.35/0.8838 & \textbf{31.42} /\textbf{0.8849} \\
         \hline
         \multirow{3}{*}{Set14} & $\times$2 & 30.24/0.8688 & 32.28/0.9056 & 32.26/0.9040 & 32.22/0.9034 & 32.42/0.9063 & 33.03/0.9124 &   \textbf{33.08} / \textbf{0.9129}   \\
         & $\times$3 & 27.55/0.7742 & 29.13/0.8188 & 29.05/0.8164 & 29.16/0.8196 & 29.28/0.8209 & 29.77/0.8314 &  \textbf{29.80}/ \textbf{0.8320}  \\
         & $\times$4 & 26.00/0.7027 & 27.32/0.7491 & 27.24/0.7451 & 27.40/0.7518 & 27.49/0.7503 & 28.01/0.7674 &   \textbf{ 28.07}/ \textbf{0.7688} \\
         \hline
         \multirow{3}{*}{BSD100} & $\times$2 & 29.56/0.8431 & 31.21/0.8863 & 31.16/0.8840 & 31.18/0.8855 & 31.36/0.8879 & \textbf{31.90}/\textbf{0.8960} &  31.89 /0.8958   \\
         &$\times$3 & 27.21/0.7385 & 28.29/0.7835 & 28.22/0.7806 & 28.29/0.7840 & 28.41/0.7863 & 28.82/0.7976 &     \textbf{28.84}/\textbf{0.7981}  \\
         &$\times$4 & 25.96/0.6675 & 26.82/0.7087 & 26.75/0.7054 & 26.84/0.7106 & 26.90/0.7101 & 27.29/0.7251 &     \textbf{27.30}/ \textbf{0.7256}  \\
         \hline
         \multirow{3}{*}{Urban100} & $\times$2 & 26.88/0.8403 & 29.20/0.8938 & 29.11/0.8904 & 29.54/0.8967 & 29.50/0.8946 & 30.76/0.9140 &     \textbf{30.82}/ \textbf{0.9144}    \\
         & $\times$3 & 24.46/0.7349 & 26.03/0.7973 & 25.86/0.7900 & 26.44/0.8088 & 26.24/0.7989 & 27.14/0.8279 & \textbf{27.17}/\textbf{0.8282} \\
         & $\times$4 & 23.14/0.6577 & 24.32/0.7183 & 24.19/0.7096 & 24.79/0.7374 & 24.52/0.7221 & 25.18/0.7524 &   \textbf{25.25}/ \textbf{0.7546}\\
         \end{tabular}
         }
   \end{table*}

%////////////////////////////////////////////////////////////////////////////////
\subsection{Importance of preserving negative information }

We conduct an ablation experiment for DPN on Set14. Table~\ref{table:ablations} shows the performance obtained with our reconstruction part which discards the ReLU layers. With a ReLU layer, the network failed to decrease the training loss or did not converge. First, all the convolutional skip connections have been changed into stacked plain layers which is a similar fashion like inception unit in Fig.~\ref{fig:Comp}(b). It broaden the width of the net. However, without the help of batch normalization, this structure failed to get good convergence with such a depth as our DPN net (40 layers) and got stuck with a high training loss and thus poor performance.

Then performance compared with convolutional skip vs. identity skip connections, and the positions of ReLU layers (ReLU before/after conv) as Fig.~\ref{fig:Comp}(c,d) respectively are represented in Table~\ref{table:ablations} for Set14. If ReLU comes after convolutional layers, the negative information will be discarded. Convolutional layers of the same depth among these networks have the same parameters. Note that in these experiments, a slightly different filter numbers of the extraction and reconstruction part has been applied so the performance is slightly different from Table~\ref{table:Proj_basic}.

  \begin{table}[thb]
  \centering
  \setlength{\tabcolsep}{10pt}
  \caption{Ablation comparisons of convolutional skip vs identity skip connections, the order of convolution and ReLU layers in terms of average PSNR (dB) on Set14.}
  \label{table:ablations}
  \begin{tabular}{c| c c c c}
      scale&conv$+$   &conv$+$ &identity$+$&identity$+$\\
		   &after\_act&pre\_act&after\_act&pre\_act\\
    \hline\hline
      $\times 2$ & \textbf{33.05}& 32.95&   32.97    & 33.01\\
      $\times 3$ & \textbf{29.78}&  29.74 &     29.75   & 29.77  \\
      $\times 4$ & \textbf{28.08}& 28.03  &    28.02   & 28.02  \\
 \end{tabular}
\end{table}

  From the results in Table~\ref{table:ablations}, we conclude that preserving negative information is effective in two ways: either by our conv projection or by rearranging the order of convolutional and ReLU layers. Interestingly, conv projection which provides a stronger representational ability combined with a non-negative description achieves best performance. The combination of the two ways of preserving negative information has not brought more benefits.

%////////////////////////////////////////////////////////////////////////////////
\subsection{Adaptation using internal contents}

As the number of internal examples is limited especially when the input image is small, small stride and scale augmentation are applied. To demonstrate that our data specific finetuning approach is a generic way to improve deep model, we finetune not only our DPN but also VDSR and compare the average PSNR (dB) results on 5 datasets for $\times 2$ in Table~\ref{table:FT_VDSR}.

  \begin{table}[!htb]
  \centering
  \caption{Average PSNR (dB) comparisons on 5 datasets for $\times 2$ of our DPN and the VDSR method with and without adaptation by finetuning to the internal contents.}
  \label{table:FT_VDSR}
  \resizebox{\linewidth}{!}
  {
  \begin{tabular}{l|r r r r r}
    %\hline
     Method                    &Set5&Set14&BSD100&Urban100 &L20\\
    \hline\hline
VDSR~\cite{kim2016accurate}&37.53&33.03&31.90&30.76&40.44\\
VDSR adapted               &37.59&33.16&31.91&30.99&40.69\\
VDSR improvement           &+0.07&+0.13&+0.01&+0.23&+0.25\\
\hline
DPN (ours)     			   &37.52&33.08&31.89&30.82&40.44\\
DPN adapted 			   &37.58&33.16&31.91&31.05&40.74\\
DPN improvement            &+0.06&+0.08&+0.02&+0.23&+0.30\\
 \end{tabular}
 }
\end{table}

\begin{figure*}[th]
  \centering
  \footnotesize
  \setlength{\tabcolsep}{1pt}
  \begin{tabular}{cccccc}
   GT& bicubic & SRCNN~\cite{dong2014learning} & VDSR~\cite{kim2016accurate} & DPN & DPN adapted\\
    \includegraphics[width=0.160\textwidth]{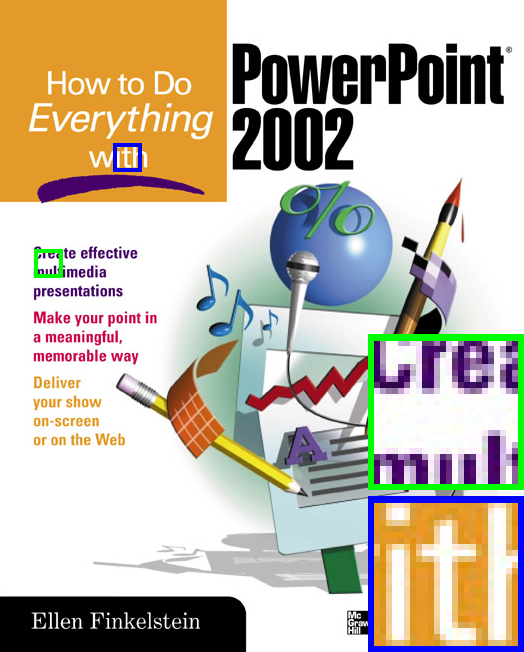}&
    \includegraphics[width=0.160\textwidth]{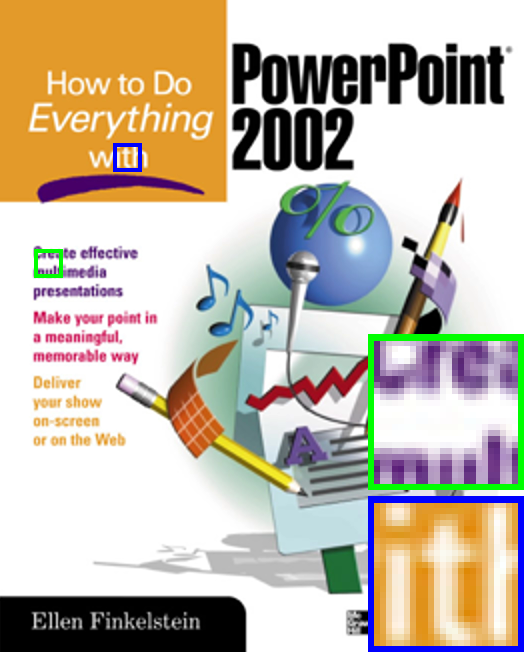}&
    \includegraphics[width=0.160\textwidth]{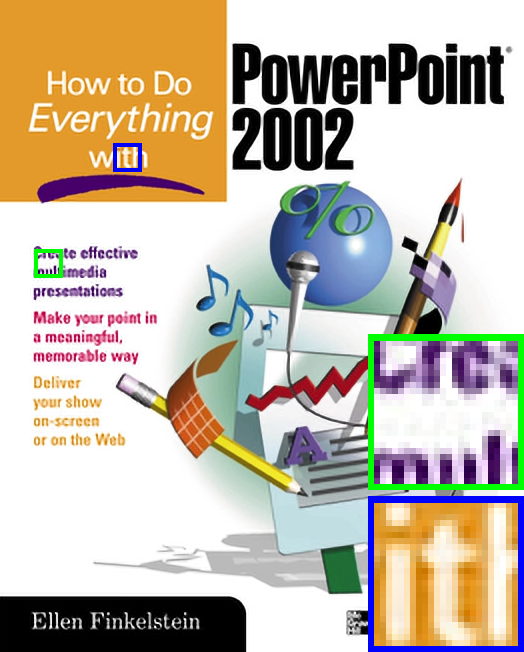}&
    \includegraphics[width=0.160\textwidth]{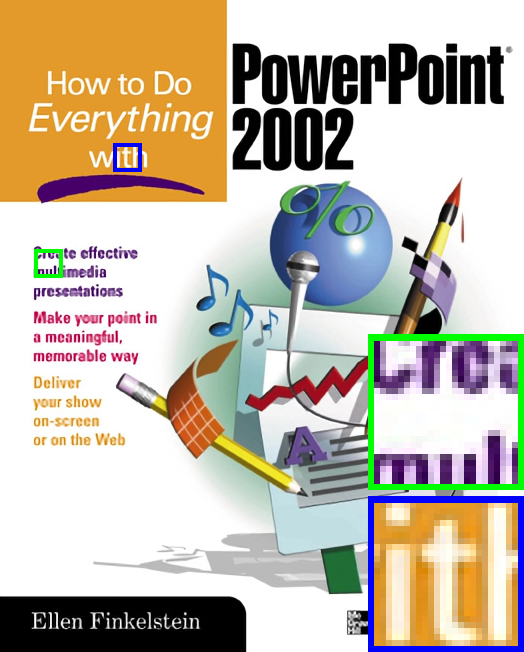} &
    \includegraphics[width=0.160\textwidth]{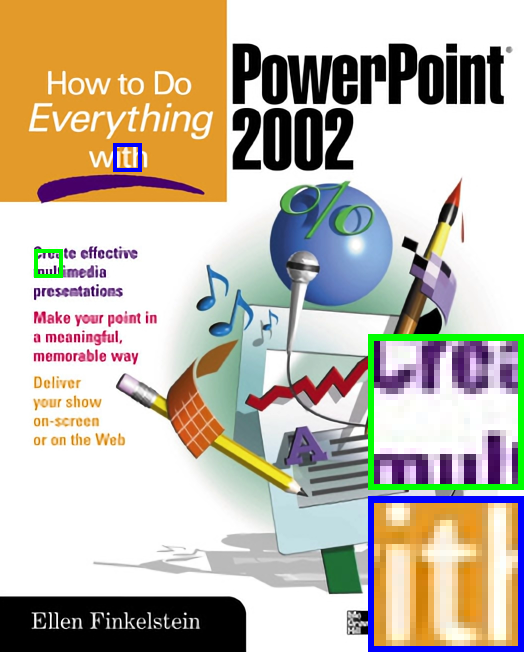}&
    \includegraphics[width=0.160\textwidth]{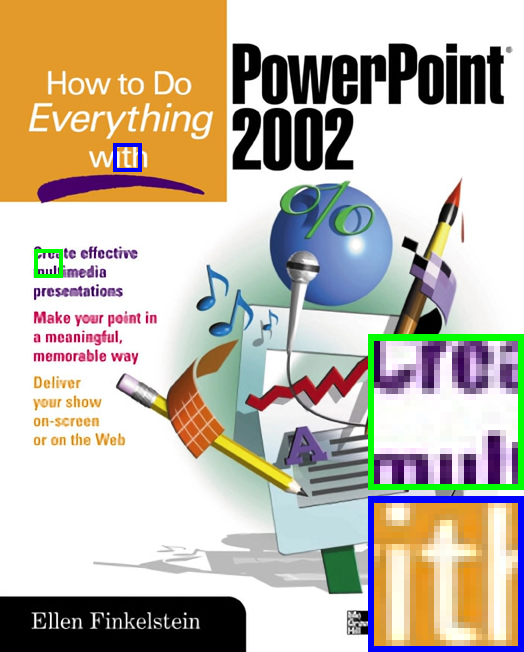} \\
PSNR / SSIM $\otimes2$ & 26.87 / 0.945 &30.53 / 0.976 &  32.93 / 0.988 &33.07 / 0.988 &  \textbf{ 33.30 / 0.989  } \\
    \includegraphics[width=0.160\textwidth]{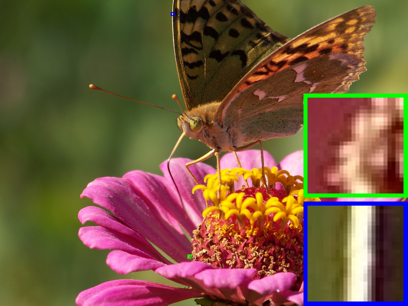}&
    \includegraphics[width=0.160\textwidth]{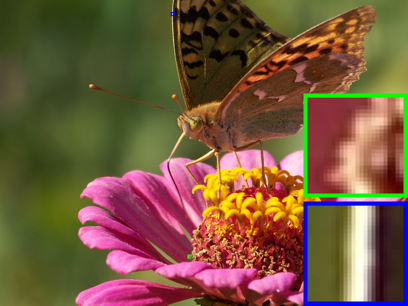}&
    \includegraphics[width=0.160\textwidth]{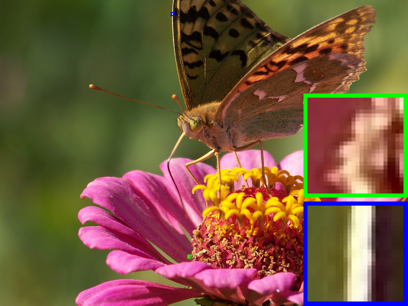}&
    \includegraphics[width=0.160\textwidth]{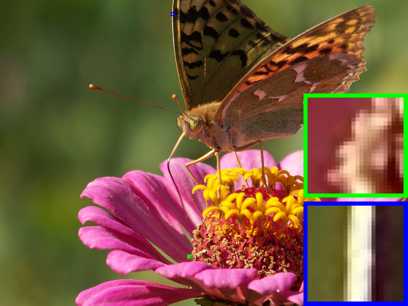} &
    \includegraphics[width=0.160\textwidth]{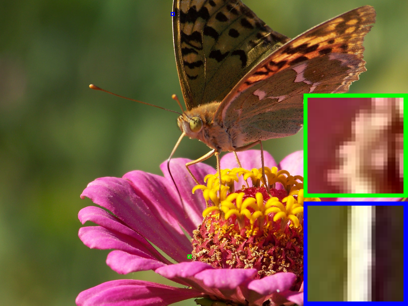}&
    \includegraphics[width=0.160\textwidth]{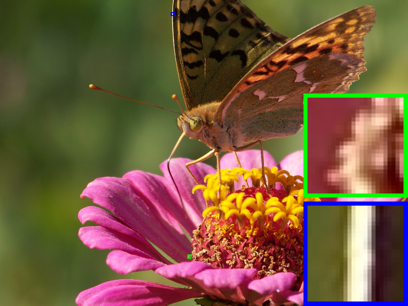} \\
PSNR / SSIM $\otimes2$& 45.13 / 0.991 & 46.80 /0.993 &  47.49 /0.994  & 47.53 / 0.994  &  \textbf{  47.95/ 0.994   } \\
    \includegraphics[width=0.160\textwidth]{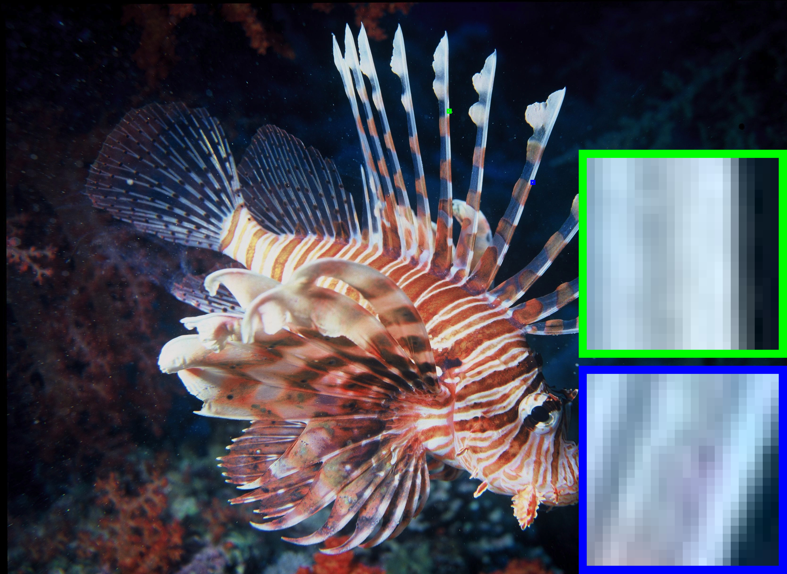}&
    \includegraphics[width=0.160\textwidth]{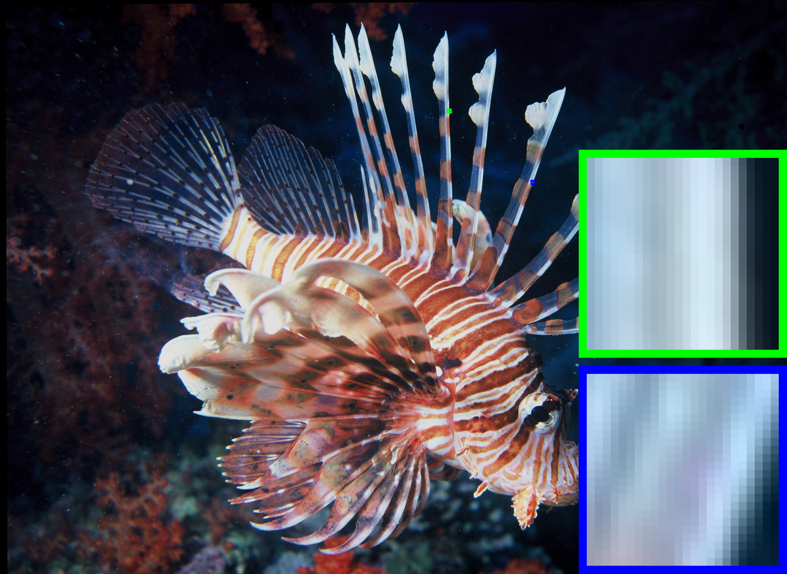}&
    \includegraphics[width=0.160\textwidth]{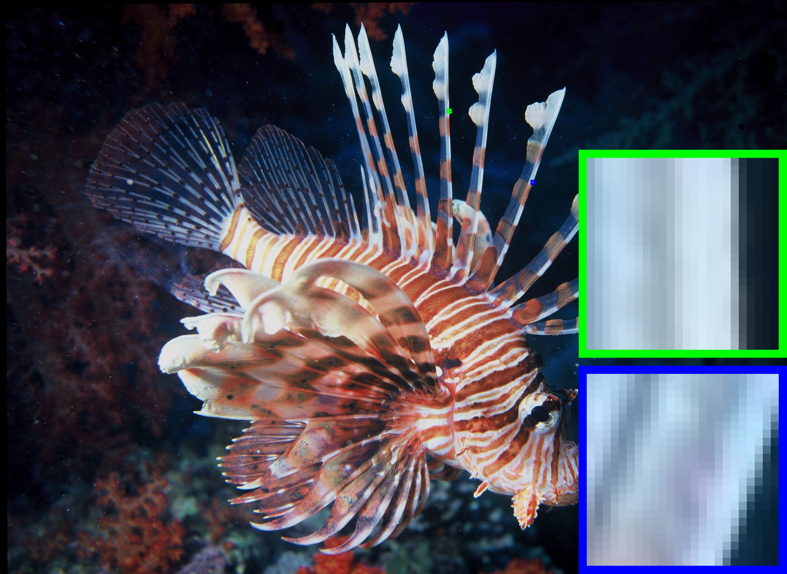}&
    \includegraphics[width=0.160\textwidth]{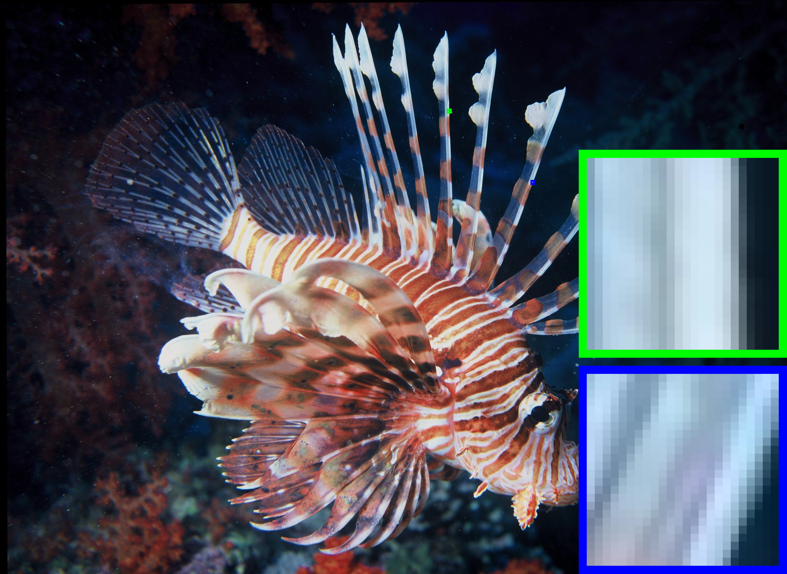} &
    \includegraphics[width=0.160\textwidth]{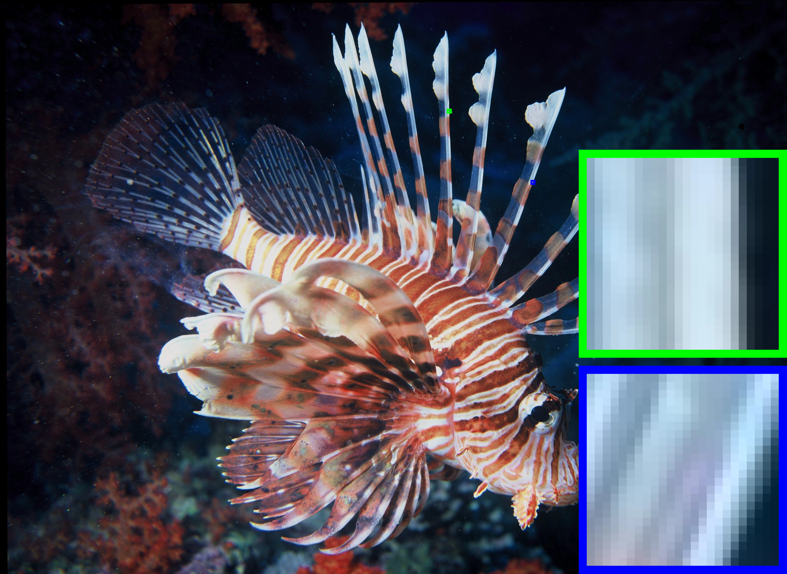}&
    \includegraphics[width=0.160\textwidth]{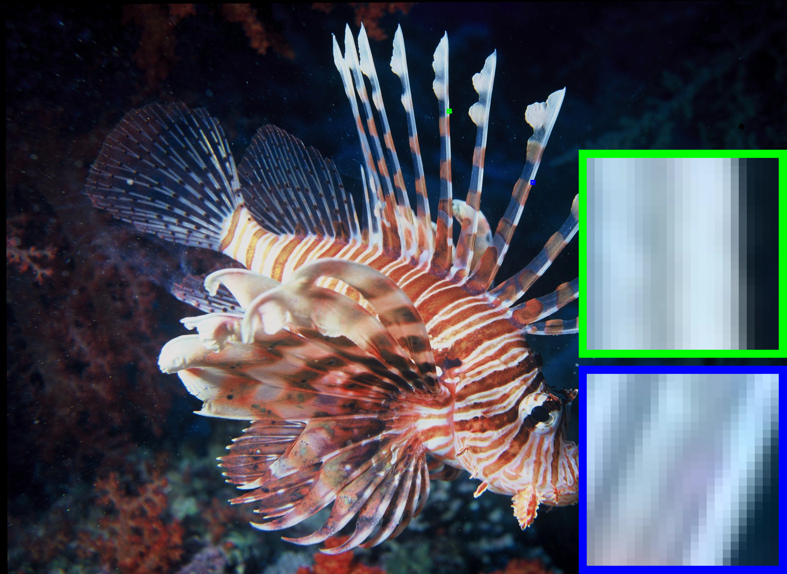} \\
PSNR / SSIM $\otimes3$& 45.13 / 0.983 & 46.19 / 0.987 &  46.57 / 0.987  &  46.77 / 0.988   &  \textbf{ 47.10  /  0.988   } \\
    \includegraphics[width=0.160\textwidth]{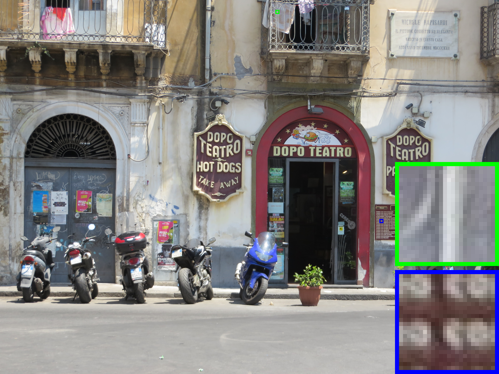}&
    \includegraphics[width=0.160\textwidth]{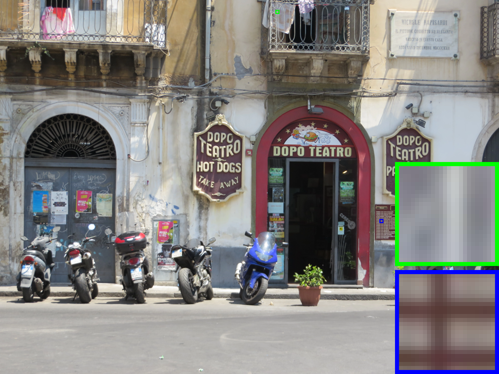}&
    \includegraphics[width=0.160\textwidth]{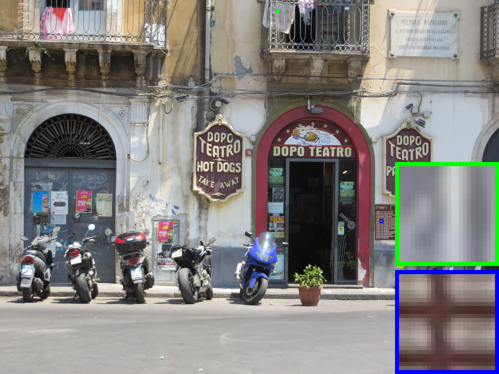}&
    \includegraphics[width=0.160\textwidth]{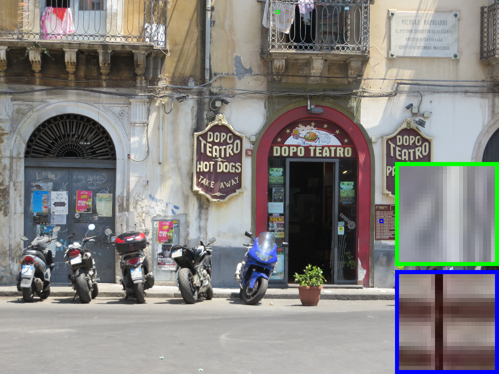} &
    \includegraphics[width=0.160\textwidth]{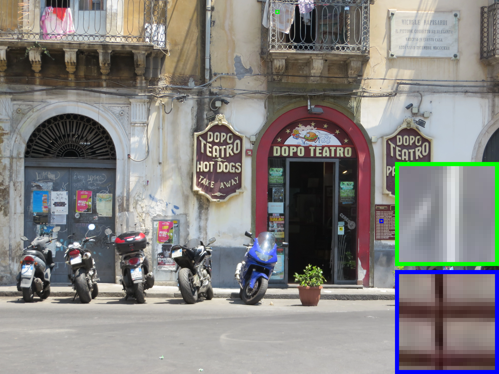}&
    \includegraphics[width=0.160\textwidth]{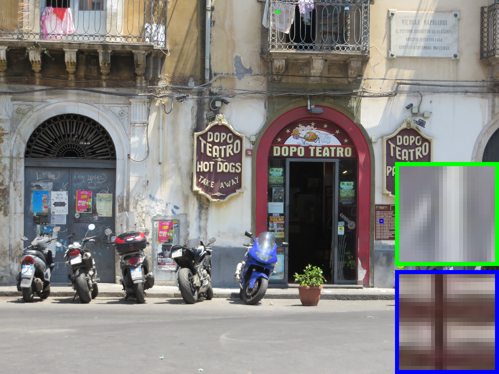} \\
PSNR / SSIM $\otimes4$& 32.52 / 0.894 &  33.89 / 0.909  &  34.66  /  0.921  &  34.71 /  0.921   &  \textbf{ 34.92   /  0.922    } \\
  \end{tabular}
  \caption{SR results for our methods -- DPN and DPN adapted by internal finetuning methods -- in comparison with VDSR~\cite{kim2016accurate}, SRCNN~\cite{dong2016image}, bicubic interpolation and ground truth image. The cropped patches are of size $25\times25$ pixels. The bottom 3 images are from L20 while the first is from Set14.}
  \label{fig:VisComp}
\end{figure*}

Not surprisingly, both DPN and VDSR largely improve after our adaptation, especially on Urban100 and L20 dataset (+0.23dB on Urban100 up to +0.3dB improvements on L20). The adapted DPN is 0.06dB better than the adapted VDSR on Urban100 and 0.05dB on L20. The visual comparisons with the state of art methods in Fig.~\ref{fig:VisComp} demonstrate that the adaptation keeps more details and sharper edges.
Urban100 has strong self similarities while L20 has large size images and therefore provide a large number of internal examples for model finetuning. The poor improvement on BSD100 is mainly due to the fact that already we used 200 BSD images at training which have a similar peculiarity with BSD100.

\subsubsection{Performance vs. image size}

\begin{figure}[!th]
  \centering
  \footnotesize
  \begin{tabular}{cc}
   \subfigure{\includegraphics[width=0.220\textwidth]{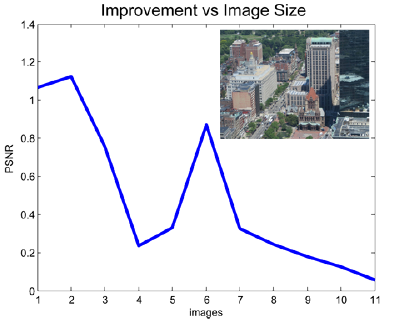}}&
  \subfigure{\includegraphics[width=0.220\textwidth]{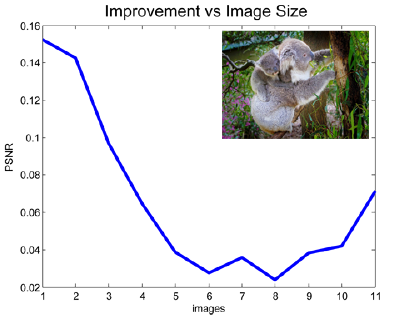}} \\
    (a) high patch redundancy & (b) low patch redundancy\\
  \end{tabular}
  \caption{PSNR (dB) improvement after DPN finetuning on each input image of different sizes. The images are ordered from the largest `1' to the smallest `11' size.}
  \label{fig:sz_img}
\end{figure}

To analyze performance vs. image size, we need first to alleviate the influence of the self-similarities/contents. For this each ground truth (GT) HR image is bicubically downsized to produce GT images of lower sizes. The obtained GT images of different sizes share self-similarities/contents among themselves and with the originating HR image. Then by downsizing we generate LR input images and record the improvements achieved on these images by our adapted by finetuning methods. Let's take two large images from L20 for an example, one with high self-similarities and another with fewer recurrent structures. The original sizes are $4000 \times 3000$ and $5184 \times 3456$. The images are downsized with a factor 1.25 per each to generated 11 GT images labeled from 1 to 11 from the largest size to the lowest. The performance improvements over the generic DPN are plotted in Fig.~\ref{fig:sz_img}.
The inflections of the curve are mainly due to the parameters changing as the image sizes become smaller than a threshold, we change the parameters (\eg image pyramid scaling factor, strides) to extract more internal examples. We tried to keep a comparable number of extracted internal examples for each image, more than 10,000 for most of the images (including augmentation).

\begin{table}[!thb]
  \centering
  \caption{Performance of adaptation by internal finetuning on L20.}
  \label{table:DPN_scales}
  \begin{tabular}{c||c c|c}
       \multirow{2}{*}{Scale} & DPN adapted & DPN & gain  \\
         & PSNR/SSIM & PSNR/SSIM &  \\
    \hline\hline
      $\times 2$ & \textbf{40.74}/\textbf{0.9655}& 40.44/0.9650& +0.3db  \\
      $\times 3$ & 36.35/0.9226 &  36.17/0.9221 &   +0.18db    \\
      $\times 4$ & 33.85/0.8809 & 33.79/0.8808  &  +0.06db    \\
 \end{tabular}
\end{table}

As expected the trend is to obtain smaller improvements with the smaller sizes of the input images which means that DPN is not able to pick sufficient internal priors for significant improvements. The same trend can be seen for the internal finetuning \wrt the magnification factor, as shown in Table~\ref{table:DPN_scales}. The larger the magnification factor / scale is the smaller the number of internal priors that can be used and, therefore, the smaller the gains are.

\subsubsection{Performance vs. finetuning epochs}

\begin{figure}[!th]
  \centering
\resizebox{\linewidth}{!}
{
  \begin{tabular}{cc}
  \subfigure{\includegraphics[width=0.22\textwidth]{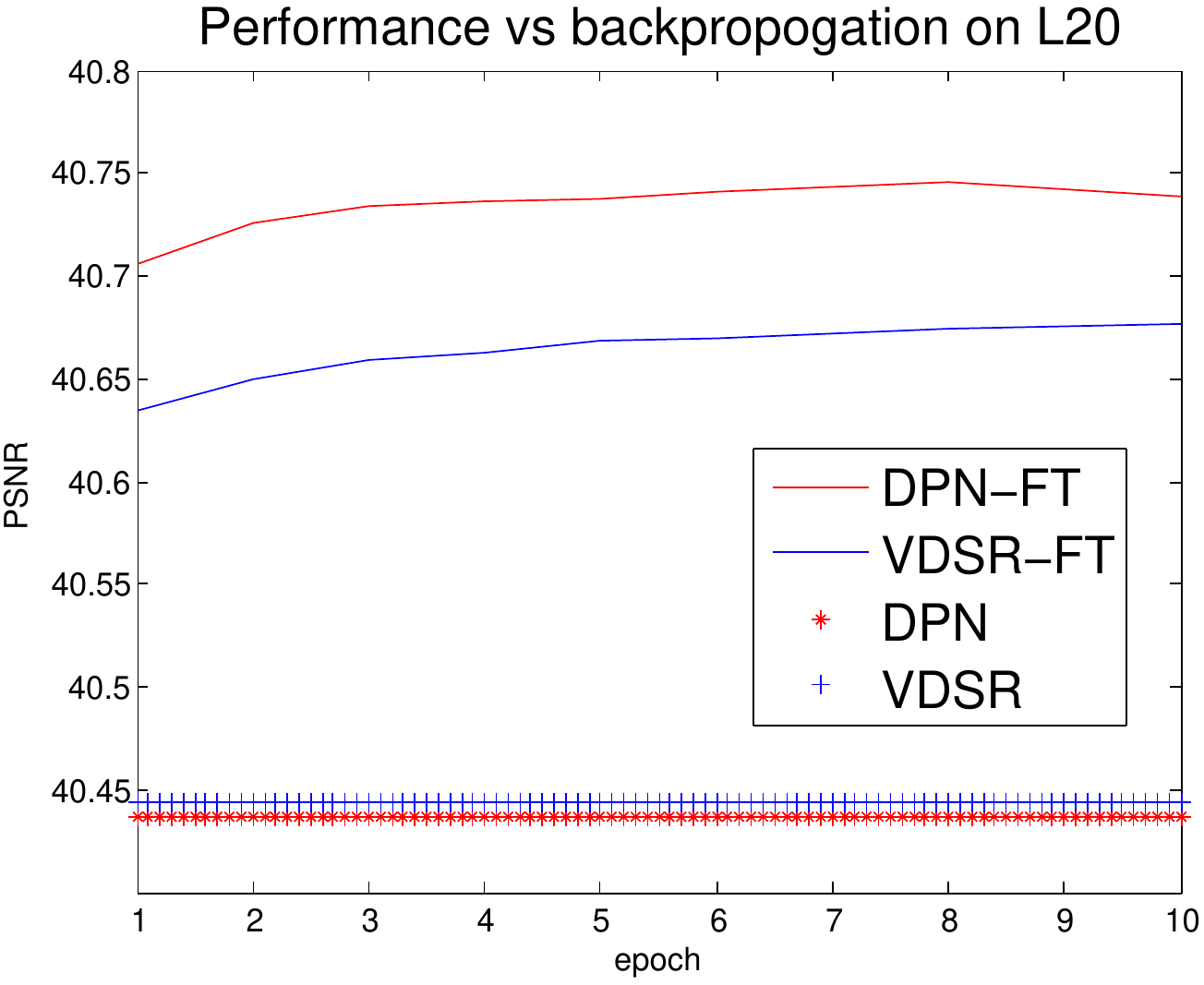}}&
  \subfigure{\includegraphics[width=0.220\textwidth]{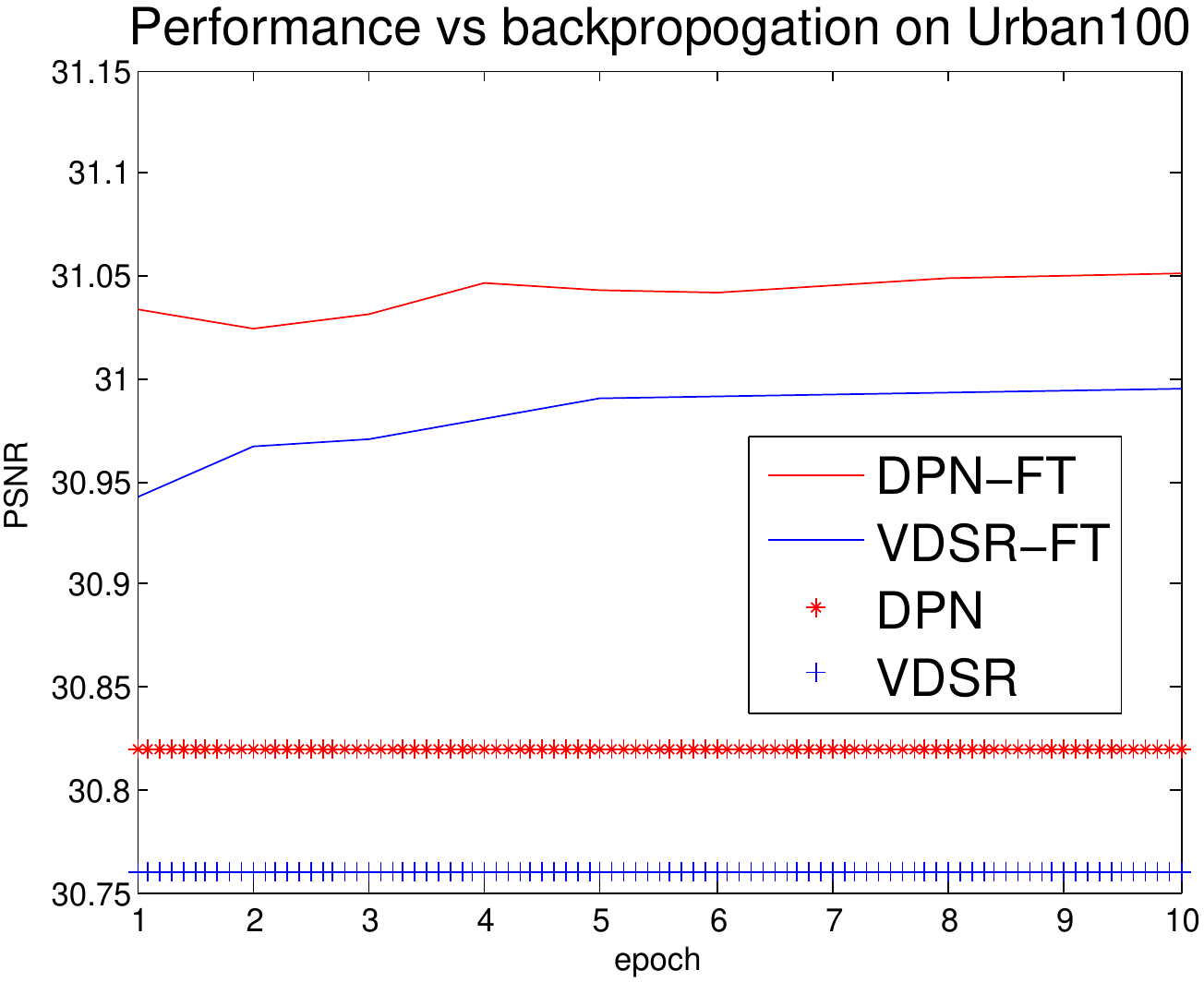}} \\
    (a) L20&(b) Urban100\\
  \end{tabular}
}
  \caption{Average PSNR (dB) performance of VDSR and our DPN method vs. finetuning epochs on L20 and Urban100 datasets in comparison with the generic externally trained VDSR and DPN methods.}
  \label{fig:bp_curve}
\end{figure}

In Fig.~\ref{fig:bp_curve} we plot the average performances over L20 and Urban100 datasets achieved by both DPN and VDSR when adapted by finetuning on internal examples from the LR test image.
We vary the number of epochs used in the finetuning and report the average performance over the whole datasets.
One single epoch (all the internal examples were backpropagated once) for adaptation makes a large difference, as most of the improvement over the baseline generic methods is achieved after a single finetuning epoch. After a couple of epochs, the improvement peaks and reaches a plateau.

\subsubsection{Runtime analysis}

Internal adaptation is a time expensive but performance superior approach. The runtime efficiency is closely related to the number of internal examples considered. When limited number of internal examples are used from the LR image (such as by loosely sampling them with a large stride or without scale augmentation which applies the input image to form a image pyramid), the adaptation in one epoch can be fast, as the input LR image is smaller.
In fact, the number of internal examples depends on the parameters applied. The performance, running time and parameters are reported in Table~\ref{table:runtime}. Note that only the stride parameter is different for Urban100 between columns and only with and without scale augmentation is different for L20.

The number of internal examples and the stride are related by $(floor((imgSz_1-subimgSz)/stride)+1) \times (floor((imgSz_2-subimgSz)/stride)+1) $. $imgSz$ and $subimgSz$ are the size of the images and examples respectively. The size For typical Urban100 images with size $1024\times 1024$ (input LR images $512 \times 512$) there are 576 samples for a stride of 20.

\begin{table*}[!htb]
 \centering
         \caption{Internal finetuning performance and runtime vs. parameter settings}
         \label{table:runtime}
\begin{tabular}{c|c|c|c|c}
  \hline
   Urban100 & DPN & stride20 & stride5 & performance in plateau \\
  (PSNR/gain/time)& 30.82/--/-- & 30.87/+0.05/7.2s & 31.04/+0.22/97.3s & 31.05/+0.23/389.2s \\
  \hline
  \hline
 L20& DPN  & single scale & scale augmentation & performance in plateau \\
  (PSNR/gain/time) & 40.44/--/-- & 40.61/+0.17/100.5s & 40.71/+0.27/288.7s & 40.74/0.30/1154.8s \\
  \hline
\end{tabular}
  \end{table*}

According to Table~\ref{table:runtime} and Fig.~\ref{fig:bp_curve}, one single epoch (all internal examples are backpropagated once) of adaptation accounts for the greatest gain over the baseline and the performance converges after a couple of epochs. Extracting internal examples for Urban100 with a stride 20 and finetune the model in one epoch (7.2 seconds) improves the performance by 0.05dB, and after 5 epochs (36.0s) by 0.13dB.

In addition, data augmentation further improves at computation cost. Since even without augmentation the performance improvement is clear, the operation can be activated whenever the application needs it. For medium size images DPN adaptation takes $\sim1$ minute but the parameters can tradeoff speed for performance.

\subsection{Adaptation using internal contents and external augmentation}
   Adaptation using internal contents and external augmentation means finetuning the model to the internal examples of the LR image plus other external examples which are similar to the internal ones. This proves to be effective when we find similar examples.

   \begin{itemize}
     \item Internal examples are extracted.
     \item External examples which has strong similarities are prepared for corresponding internal examples.
     \item Mixing internal examples and corresponding external examples together.
     \item Finetuning the model using these mixtures of internal contents and external augmentation.
   \end{itemize}

   In the following we give an example of how external augmentation can help when we find (very) similar examples. The ground truth images of Set14 were downscaled by a factor of 1.25 and used as external images. Then with a mixing of external and internal examples, the performance boosted as shown in Table~\ref{table:DPN_aug} denoted by `adapted DPN+aug(images)'. This experiment demonstrated that with an proper (ideal) external example augmentation, the adaptation using internal contents and external augmentation can work. With a fast image retrieval technique this strategy could be very effective even for very small images.
    Moreover, if a large and diverse external example set is prepared and each internal example could find some external examples which have strong similarities by by some efficient and effective nearest search methods or matching methods like PatchMatch~\cite{barnes2009patchmatch}. Larger improvements are expected and this strategy will be explored in the future.

   \begin{table}[!htb]
  \centering
  \caption{DPN performance comparison with and without adaptation to internal contents and external augmentation on Set14 with magnification $\times2$.}
  \label{table:DPN_aug}
\begin{tabular}{c|c|c|c}
    \hline
       &DPN&adapted DPN & adapted DPN+aug(images)\\
    \hline
   PSNR (dB)& 33.08 & 33.16 &33.29\\
    \hline
 \end{tabular}

\end{table}

\subsection{Adaptation as a model selection}
In the next we described a couple of experiments involving the selection of models from a pool of trained models.

\subsubsection{Experiment 1}
We used the adaptation using internal contents as described in Section~\ref{sec:adap_select} to finetune 20 DPN models corresponding to the L20 images in addition to the original DPN model and test them on Set14 images with $\times 2$. By choosing the best 3 models out of 20 for restoration, as reflected by the performance on the downscaled LR images, the results slightly improve to 33.12dB \wrt the generic VDSR with 33.03dB. Although it is a marginal improvement, partly due to the large mismatch between the L20 and Set14 images, this strategy has a great potential. Various strategies can be devised to train a set of SR models specialized on (semantically) different images and image contexts (such as cartoons, flowers, skyscrapers, villages, dogs).

 \begin{table}[!htb]
  \centering
  \caption{a successful model selection example}
  \label{table:adaptbyanother}
  \begin{tabular}{c||c|c||c|c}
           & \multicolumn{4}{c}{model}\\
           \cline{2-5}
           & \multicolumn{2}{c||}{VDSR}&\multicolumn{2}{c}{DPN}\\
      \hline
    \backslashbox{\scriptsize image tested on}{\scriptsize image adapted to}  &img47&img72&img47&img72\\
    \hline\hline
     img47&26.52&26.29&26.69&26.22\\
    \hline
      img72&25.52&26.13&25.60&26.40\\
 \end{tabular}
\end{table}

\subsubsection{Experiment 2}
\begin{figure*}
\setlength{\tabcolsep}{1pt}
\centering
\resizebox{\linewidth}{!}
{
\begin{tabular}{cccccc}
LR input & bicubic & VDSR~\cite{kim2016accurate} & DPN & DPN (adapted to 47) & DPN (adapted to 72)\\
47
\includegraphics[width=0.1\linewidth]{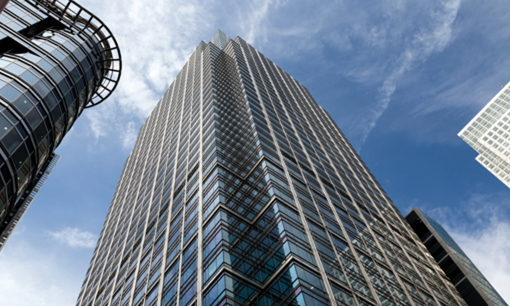}&
\includegraphics[width=0.25\linewidth]{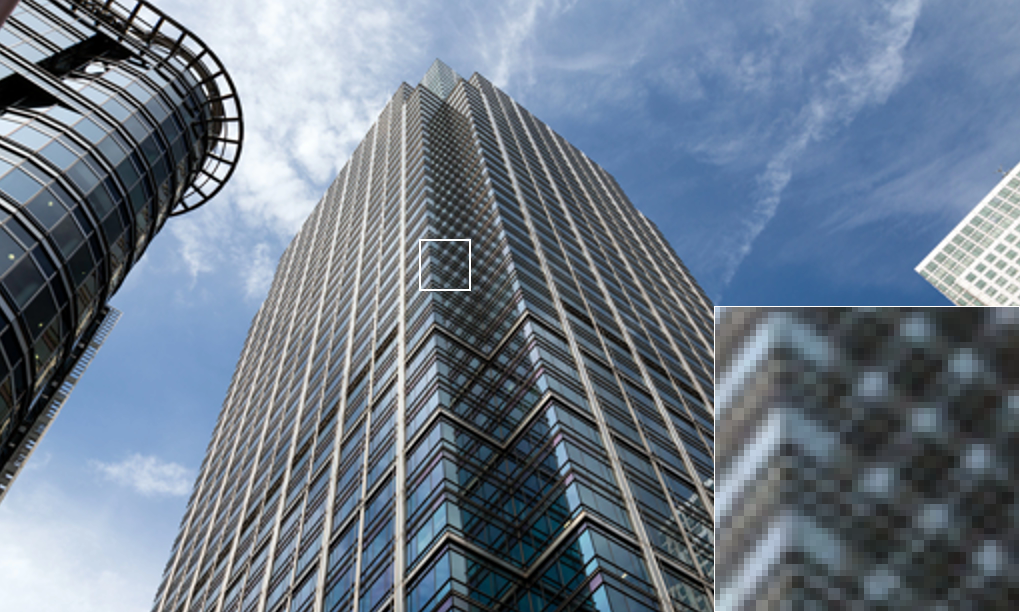}&
\includegraphics[width=0.25\linewidth]{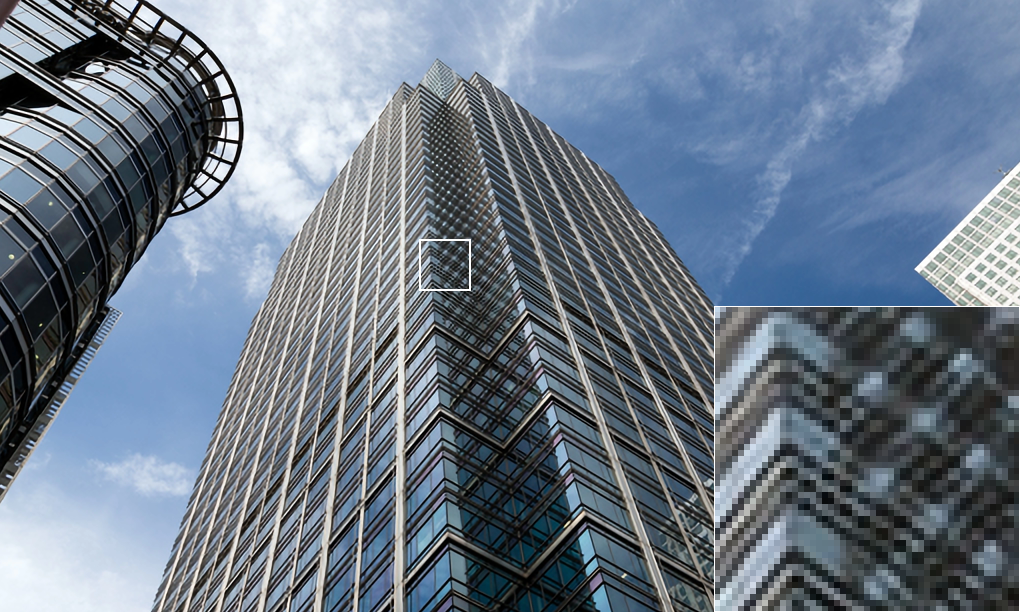}&
\includegraphics[width=0.25\linewidth]{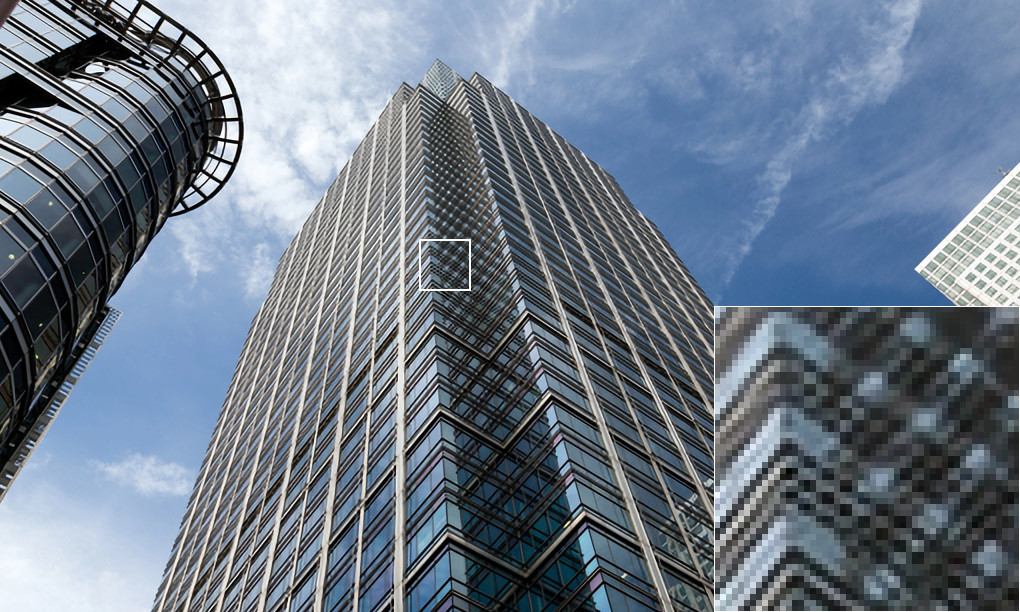}&
\includegraphics[width=0.25\linewidth]{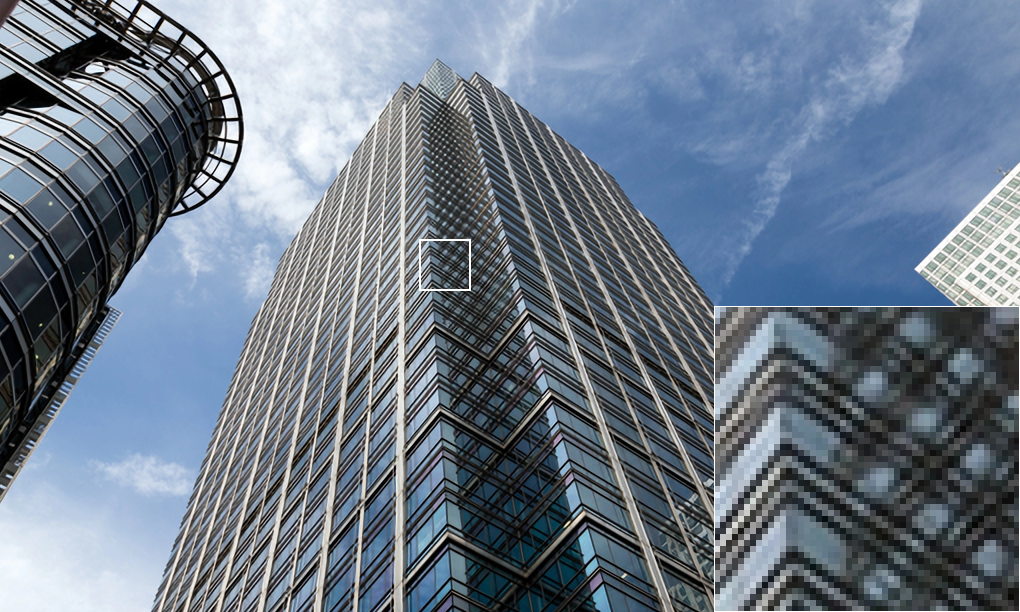}&
\includegraphics[width=0.25\linewidth]{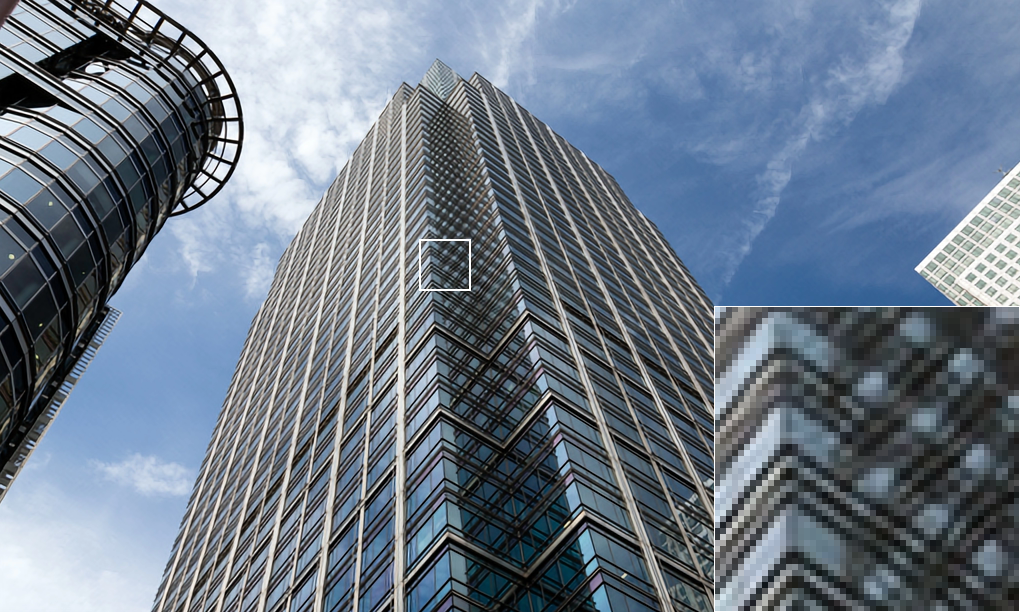}\\
PSNR / SSIM&22.95 / 0.821&25.99 / 0.918 & 26.04 / 0.919&\textbf{26.69 / 0.929} & 26.22 / 0.919\\
72
\includegraphics[width=0.1\linewidth]{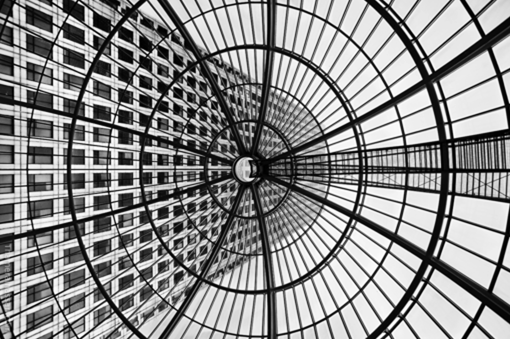}&
\includegraphics[width=0.25\linewidth]{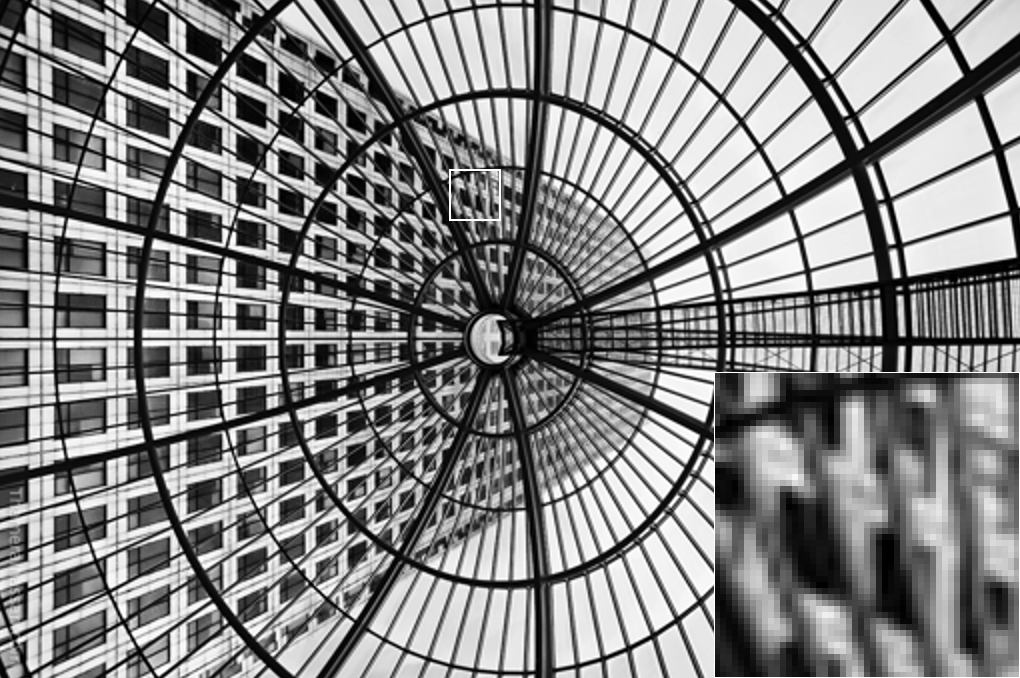}&
\includegraphics[width=0.25\linewidth]{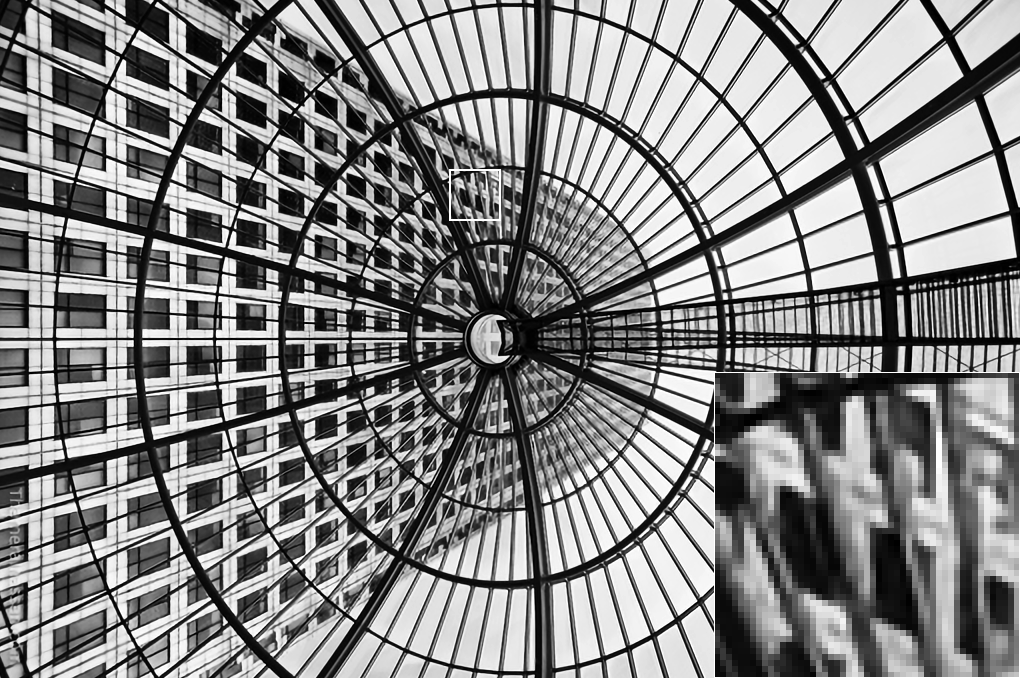}&
\includegraphics[width=0.25\linewidth]{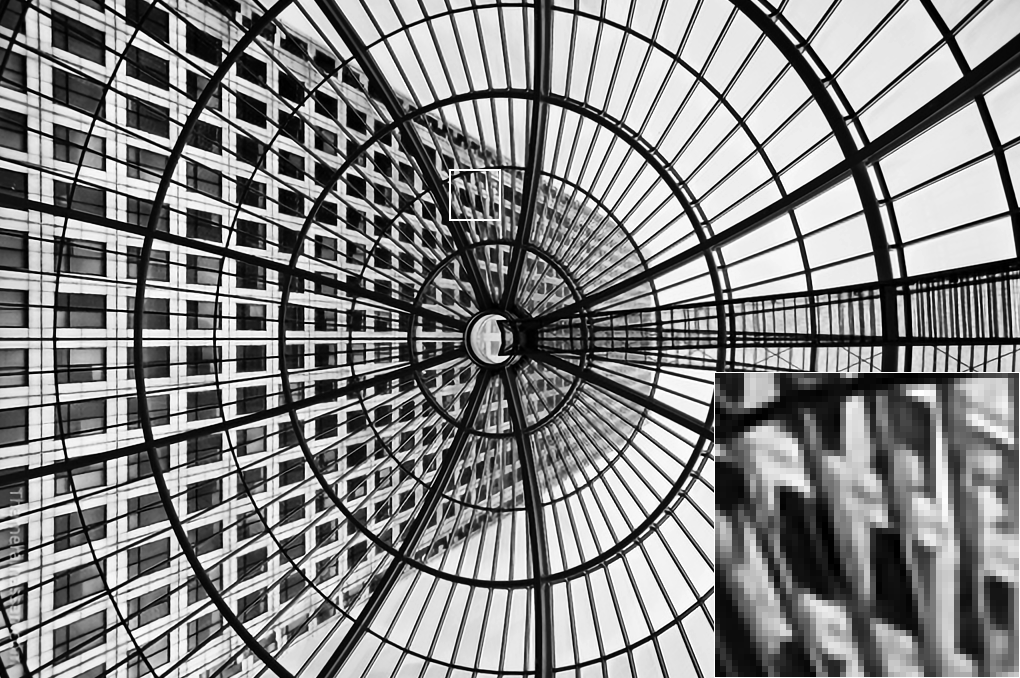}&
\includegraphics[width=0.25\linewidth]{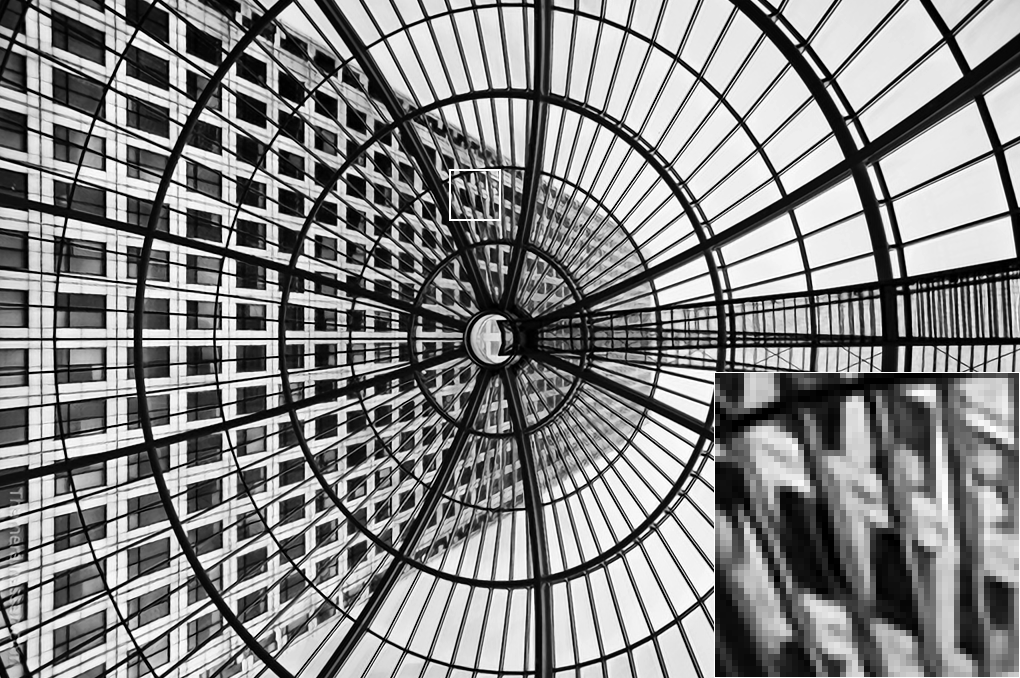}&
\includegraphics[width=0.25\linewidth]{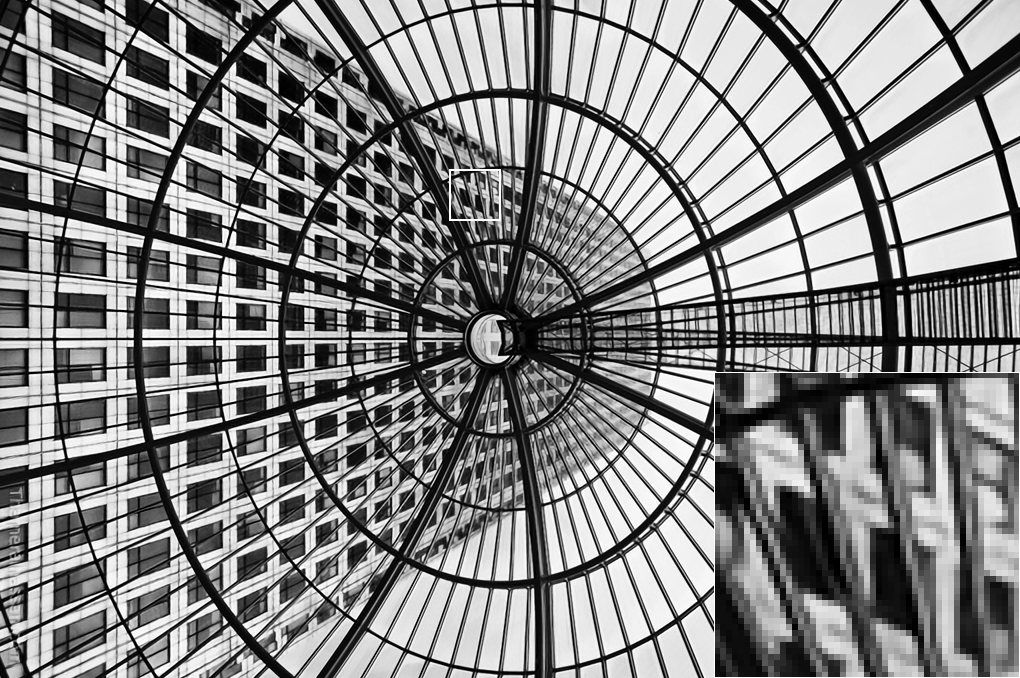}\\
PSNR / SSIM&20.39 / 0.826&25.11 / 0.944 & 25.37 / 0.946&25.60 / 0.946 & \textbf{26.40 / 0.953}\\
\end{tabular}
}
\caption{SR $\times 2$ results on Urban100 images 47 and 72 whose finetuned models (on each LR image) work well for each other.}
  \label{fig:urban47_72}
\end{figure*}

In a second experiment, we take two images from Urban100, image 47 and image 72. In contrast to the previous experiment the two images share similar contents and, in this case, their finetuned VDSR / DPN models work very well for each other. As shown Table~\ref{table:adaptbyanother}, testing image 47, with the VDSR model finetuned on image 72 brings an improvement of 0.41dB over the generic VDSR, while on the counterpart, testing image 72, with the VDSR model finetuned on image 47 shows an improvement of 0.30dB. On the same images our DPN works better than VDSR improving 0.05dB on image 47 and 0.26dB on image 72 and when using the finetuned DPN models we see the same behavior as for VDSR. The images 47 and 72 and the SR results of our DPN method are found in Fig~\ref{fig:urban47_72}.
These experimental results show the potential of model selection if the pool of models are large and diverse enough.

\subsubsection{Experiment 3}
The performance of a model for the `input restoration' and `HR image super-resolution' is highly correlated as shown in Section~\ref{sec:adap_select}.

In Table~\ref{table:Model2Ada} for Urban100 dataset and $\times2$ we compare our DPN with DPN with internal adaptation (DPN(A)), and given 100 DPN models built on the low resolution Urban images, our selection strategy based on the LR input image and its downscaled version (Selection), the oracle selection of the best model (Selection(O)), and our selection strategy from a random subset of 10 models (Random Selection(10)). Our selection strategy picks the best model according to the performance of restoring LR input image from downscaled LR images. The Table~\ref{table:Model2Ada} demonstrates that our model selection strategy is effective as an alternative way for model adaptation by finetuning, both achieving +0.23dB improvements. For 80 Urban images the model selection strategy led to better performance than the DPN generic model.
Noteworthy is that that a random selection strategy harms the performance while the optimal /oracle selection strategy leads to only 0.14dB better performance than the proposed strategy.

\begin{table}[!hb]
  \centering
  \caption{Model selection vs internal adaptation on Urban100 and $\times2$}
  \label{table:Model2Ada}
\resizebox{\linewidth}{!}
{
  \begin{tabular}{c||c|c|c|c|c}
              Method & DPN & DPN & Selection &Selection& Random\\
                     &     & (A) & (ours)    & (O) & Selection(10)\\
               \hline\hline
              PSNR   & 30.82 & 31.05 & 31.05 &31.19 & 30.29\\
              gain   & --& +0.23 &+0.23&+0.37&-0.53\\
            \end{tabular}
}
\end{table}

 \section{Model adaptation with internal examples vs. back-projection and enhancement}
    Our model adaptation with internal examples mainly depends on the self-similarities of the images.
    We compare the (A)adaptation to the LR image by use of internal examples with two other techniques known to further improve the performance of a SR method: (B)back-projection~\cite{irani1991improving} and (E)enhanced prediction~\cite{timofte2016seven}.
As in \cite{timofte2016seven}, iterative back projection (B) refinement makes the HR reconstruction consistent with the LR input. (E) enhancement means to enhanced the restoration by averaging the predictions on a set of transformed images derived from it. For SR image, after rotations and flips, the same HR results at pixel level should be generated. Therefore, a set of 8 LR images were generated as we apply rotations and flips on the LR image as \cite{timofte2016seven}, then apply the SR method on each, reverse the transformation on the HR outputs and average for the final HR result.

\begin{table*}[!tbh]
  \centering
  \caption{Performance comparisons between model adaptation with internal examples (A), back-projection (B) and enhanced prediction (E) for magnification $\times2$. PSNR and gain are measured in dB.}
  \label{table:comparison}
  \begin{tabular}{c|c|c|c|c|c|c}
        & DPN & DPN+(A) & DPN+(B)& DPN+(E) &DPN+(A+B)&DPN+(A+E) \\
        & PSNR/SSIM & PSNR/SSIM &PSNR/SSIM& PSNR/SSIM&PSNR/SSIM&PSNR/SSIM \\
   \hline
   \hline
    Set5&    37.52 / 0.9586& 37.58 / 0.9588 &37.56 / 0.9589 &37.64	/ 0.9592 & 37.64 / 0.9591 & 37.70 / 0.9594 \\
   gain vs. DPN & --& 0.06 & 0.04 & 0.12 & 0.12& 0.18\\
   \hline
      Set14& 33.08 / 0.9129& 33.16 / 0.9132 & 33.11	/ 0.9133& 33.18 / 0.9138& 33.20 / 0.9136& 33.27/ 0.9143\\
        gain vs. DPN &-- & 0.08&0.03 & 0.10&0.12 &0.19 \\
      \hline
      BSD100& 31.89 /0.8958& 31.91 / 0.896& 31.91 /	0.8961&31.96 / 0.8967 & 31.93	/ 0.8963& 31.97 /0.8967\\
        gain vs. DPN &-- &0.02 &0.02 &0.07 &0.04 & 0.08\\
      \hline
     Urban100& 30.82 / 0.9144 & 31.05 / 0.9161& 30.85 /	0.9148& 30.95 /	0.9159 & 31.11 / 0.9167&  31.21 / 0.9179\\
       gain vs. DPN &-- & 0.23&0.03 & 0.13& 0.29&0.39 \\
      \hline
     L20& 40.44 / 0.965& 40.74 / 0.9655& 40.49 / 0.9653 & 40.56 / 0.9655 &40.78 /	0.9657  & 40.84 / 0.9659  \\
      gain vs. DPN &--&0.30 &0.05 &0.12 &0.34 &0.40 \\
 \end{tabular}
\end{table*}

  From the Table \ref{table:comparison}, some conclusions can be obtained:
    \begin{itemize}
      \item Model adaptation (A) with internal examples, back-projections(B) and enhancement(E) are all effective way to improve performance.
      \item For images with redundant contents/ high self-similarities as in L20 or Urban100, model adaptation (A) with internal examples which is especially common in larger and larger nowadays images, is more effective than the other techniques.
      \item The combination of techniques bring larger gains which indicates complementarity. DPN+(A+E) gain +0.4dB over DPN on both Urban100 and L20 while DPN+(E) gets only +0.1dB.
    \end{itemize}
    Thus we hope to arouse the interests of the communities for the internal example priors adaptation which can be combined with other techniques to further improve the situations.
\section{Discussion}
\label{sec:discussion}
The large improvements achieved by model adaptation to the input image demonstrates the power of internal example priors. Also, it is an indicator that the representation ability of a deep CNN with certain parameters is far from its limit. Our \textbf{DPN} starts from architecture design and external example prior modeling. It benefits from reducing the training difficulties and increasing the representational ability with a projection skip connection as well as information preserving reconstruction design. Moreover, when redundant contents / high self-similarities exist, which is especially common in larger and larger nowadays images, the super-resolution results can be largely improved by simple finetuning adaption of the deep model to internal examples to the input image. A model selection strategy is proposed which is proved to be efficient and effective if the pool of models are large and diverse enough. In addition, when internal prior is combined with (B) back-projection and/or (E) enhanced prediction, it leads to extra performance gains which indicates the complementarity of these techniques. To the best of our knowledge our work is the first to study adaptation of deep models for single-image super-resolution.
We want to arouse the interests of communities to focus on the internal priors which are limited but has been proved effective, and highly relevant. In the future, deep architectures to directly model internal priors will be explored.

\section{Conclusion}
\label{sec:conclusion}
We proposed a novel deep architecture for single image super-resolution and studied a couple of ways to adapt the externally learned deep models to the content of the low-resolution image input. With our proposed method we are able to improve over the state-of-the-art deep learning method (VDSR) on standard benchmarks. The adaptation to the content proved successful in most of the cases, significant improvements being achieved especially for medium to large images with redundant contents / high self-similarities.

\appendices

\section*{Acknowledgment}
This work is partially supported by National Science Foundation of China under Grant NO. 61473219, and the National Basic Research Program of China (973 Program) under Grant No. 2015CB351705.

\ifCLASSOPTIONcaptionsoff
  \newpage
\fi

{\small
\bibliographystyle{IEEEtran}
\bibliography{ImgSR}

\begin{thebibliography}{10}
\providecommand{\url}[1]{#1}
\csname url@samestyle\endcsname
\providecommand{\newblock}{\relax}
\providecommand{\bibinfo}[2]{#2}
\providecommand{\BIBentrySTDinterwordspacing}{\spaceskip=0pt\relax}
\providecommand{\BIBentryALTinterwordstretchfactor}{4}
\providecommand{\BIBentryALTinterwordspacing}{\spaceskip=\fontdimen2\font plus
\BIBentryALTinterwordstretchfactor\fontdimen3\font minus
  \fontdimen4\font\relax}
\providecommand{\BIBforeignlanguage}[2]{{%
\expandafter\ifx\csname l@#1\endcsname\relax
\typeout{** WARNING: IEEEtran.bst: No hyphenation pattern has been}%
\typeout{** loaded for the language `#1'. Using the pattern for}%
\typeout{** the default language instead.}%
\else
\language=\csname l@#1\endcsname
\fi
#2}}
\providecommand{\BIBdecl}{\relax}
\BIBdecl

\bibitem{keys1981cubic}
R.~Keys, ``Cubic convolution interpolation for digital image processing,''
  \emph{Acoustics, Speech and Signal Processing, IEEE Transactions on},
  vol.~29, no.~6, pp. 1153--1160, 1981.

\bibitem{irani1993motion}
M.~Irani and S.~Peleg, ``Motion analysis for image enhancement: Resolution,
  occlusion, and transparency,'' \emph{Journal of Visual Communication and
  Image Representation}, vol.~4, no.~4, pp. 324--335, 1993.

\bibitem{aly2005image}
H.~A. Aly and E.~Dubois, ``Image up-sampling using total-variation
  regularization with a new observation model,'' \emph{Image Processing, IEEE
  Transactions on}, vol.~14, no.~10, pp. 1647--1659, 2005.

\bibitem{freeman2000learning}
W.~T. Freeman, E.~C. Pasztor, and O.~T. Carmichael, ``Learning low-level
  vision,'' \emph{International journal of computer vision}, vol.~40, no.~1,
  pp. 25--47, 2000.

\bibitem{yang2008image}
J.~Yang, J.~Wright, T.~Huang, and Y.~Ma, ``Image super-resolution as sparse
  representation of raw image patches,'' in \emph{Computer Vision and Pattern
  Recognition, 2008. CVPR 2008. IEEE Conference on}.\hskip 1em plus 0.5em minus
  0.4em\relax IEEE, 2008, pp. 1--8.

\bibitem{timofte2013anchored}
R.~Timofte, V.~De, and L.~V. Gool, ``Anchored neighborhood regression for fast
  example-based super-resolution,'' in \emph{Computer Vision (ICCV), 2013 IEEE
  International Conference on}.\hskip 1em plus 0.5em minus 0.4em\relax IEEE,
  2013, pp. 1920--1927.

\bibitem{glasner2009super}
D.~Glasner, S.~Bagon, and M.~Irani, ``Super-resolution from a single image,''
  in \emph{Computer Vision, 2009 IEEE 12th International Conference on}.\hskip
  1em plus 0.5em minus 0.4em\relax IEEE, 2009, pp. 349--356.

\bibitem{freedman2011image}
G.~Freedman and R.~Fattal, ``Image and video upscaling from local
  self-examples,'' \emph{ACM Transactions on Graphics (TOG)}, vol.~30, no.~2,
  p.~12, 2011.

\bibitem{zontak2011internal}
M.~Zontak and M.~Irani, ``Internal statistics of a single natural image,'' in
  \emph{Computer Vision and Pattern Recognition (CVPR), 2011 IEEE Conference
  on}.\hskip 1em plus 0.5em minus 0.4em\relax IEEE, 2011, pp. 977--984.

\bibitem{chang2004super}
H.~Chang, D.-Y. Yeung, and Y.~Xiong, ``Super-resolution through neighbor
  embedding,'' in \emph{Computer Vision and Pattern Recognition, 2004. CVPR
  2004. Proceedings of the 2004 IEEE Computer Society Conference on},
  vol.~1.\hskip 1em plus 0.5em minus 0.4em\relax IEEE, 2004, pp. I--275.

\bibitem{dong2016image}
C.~Dong, C.~C. Loy, K.~He, and X.~Tang, ``Image super-resolution using deep
  convolutional networks,'' \emph{IEEE transactions on pattern analysis and
  machine intelligence}, vol.~38, no.~2, pp. 295--307, 2016.

\bibitem{yang2013fast}
J.~Yang, Z.~Lin, and S.~Cohen, ``Fast image super-resolution based on in-place
  example regression,'' in \emph{Computer Vision and Pattern Recognition
  (CVPR), 2013 IEEE Conference on}.\hskip 1em plus 0.5em minus 0.4em\relax
  IEEE, 2013, pp. 1059--1066.

\bibitem{timofte2016seven}
R.~Timofte, R.~Rothe, and L.~Van~Gool, ``Seven ways to improve example-based
  single image super resolution,'' in \emph{Proceedings of the IEEE Conference
  on Computer Vision and Pattern Recognition}, 2016, pp. 1865--1873.

\bibitem{kim2016accurate}
J.~Kim, J.~K. Lee, and K.~M. Lee, ``Accurate image super-resolution using very
  deep convolutional networks,'' in \emph{Proceedings of the IEEE Conference on
  Computer Vision and Pattern Recognition}, 2016.

\bibitem{dong2014learning}
C.~Dong, C.~C. Loy, K.~He, and X.~Tang, ``Learning a deep convolutional network
  for image super-resolution,'' in \emph{Computer Vision--ECCV 2014}.\hskip 1em
  plus 0.5em minus 0.4em\relax Springer, 2014, pp. 184--199.

\bibitem{burger2013learning}
H.~C. Burger, C.~Schuler, and S.~Harmeling, ``Learning how to combine internal
  and external denoising methods,'' in \emph{German Conference on Pattern
  Recognition}.\hskip 1em plus 0.5em minus 0.4em\relax Springer, 2013, pp.
  121--130.

\bibitem{dong2013nonlocally}
W.~Dong, L.~Zhang, G.~Shi, and X.~Li, ``Nonlocally centralized sparse
  representation for image restoration,'' \emph{IEEE Transactions on Image
  Processing}, vol.~22, no.~4, pp. 1620--1630, 2013.

\bibitem{timofte2014a+}
R.~Timofte, V.~De~Smet, and L.~Van~Gool, ``A+: Adjusted anchored neighborhood
  regression for fast super-resolution,'' in \emph{Computer Vision--ACCV
  2014}.\hskip 1em plus 0.5em minus 0.4em\relax Springer, 2014, pp. 111--126.

\bibitem{he2015deep}
K.~He, X.~Zhang, S.~Ren, and J.~Sun, ``Deep residual learning for image
  recognition,'' in \emph{Proceedings of the IEEE Conference on Computer Vision
  and Pattern Recognition}, 2016, pp. 770--778.

\bibitem{simonyan2014very}
K.~Simonyan and A.~Zisserman, ``Very deep convolutional networks for
  large-scale image recognition,'' \emph{arXiv preprint arXiv:1409.1556}, 2014.

\bibitem{he2016identity}
K.~He, X.~Zhang, S.~Ren, and J.~Sun, ``Identity mappings in deep residual
  networks,'' in \emph{European Conference on Computer Vision}.\hskip 1em plus
  0.5em minus 0.4em\relax Springer, 2016, pp. 630--645.

\bibitem{dong2016accelerating}
C.~Dong, C.~C. Loy, and X.~Tang, ``Accelerating the super-resolution
  convolutional neural network,'' in \emph{European Conference on Computer
  Vision}.\hskip 1em plus 0.5em minus 0.4em\relax Springer, 2016, pp. 391--407.

\bibitem{arXiv:1412.4564}
A.~Vedaldi and K.~Lenc, ``Matconvnet -- convolutional neural networks for
  matlab,'' \emph{CoRR}, vol. abs/1412.4564, 2014.

\bibitem{schulter2015fast}
S.~Schulter, C.~Leistner, and H.~Bischof, ``Fast and accurate image upscaling
  with super-resolution forests,'' in \emph{Proceedings of the IEEE Conference
  on Computer Vision and Pattern Recognition}, 2015, pp. 3791--3799.

\bibitem{martin2001database}
D.~Martin, C.~Fowlkes, D.~Tal, and J.~Malik, ``A database of human segmented
  natural images and its application to evaluating segmentation algorithms and
  measuring ecological statistics,'' in \emph{Computer Vision, 2001. ICCV 2001.
  Proceedings. Eighth IEEE International Conference on}, vol.~2.\hskip 1em plus
  0.5em minus 0.4em\relax IEEE, 2001, pp. 416--423.

\bibitem{huang2015single}
J.-B. Huang, A.~Singh, and N.~Ahuja, ``Single image super-resolution from
  transformed self-exemplars,'' in \emph{2015 IEEE Conference on Computer
  Vision and Pattern Recognition (CVPR)}.\hskip 1em plus 0.5em minus
  0.4em\relax IEEE, 2015, pp. 5197--5206.

\bibitem{yang2014singleBenchmark}
C.-Y. Yang, C.~Ma, and M.-H. Yang, ``Single-image super-resolution: A
  benchmark,'' in \emph{Computer Vision--ECCV 2014}.\hskip 1em plus 0.5em minus
  0.4em\relax Springer, 2014, pp. 372--386.

\bibitem{wang2015deep}
Z.~Wang, D.~Liu, J.~Yang, W.~Han, and T.~Huang, ``Deep networks for image
  super-resolution with sparse prior,'' in \emph{Proceedings of the IEEE
  International Conference on Computer Vision}, 2015, pp. 370--378.

\bibitem{barnes2009patchmatch}
C.~Barnes, E.~Shechtman, A.~Finkelstein, and D.~Goldman, ``Patchmatch: a
  randomized correspondence algorithm for structural image editing,'' \emph{ACM
  Transactions on Graphics-TOG}, vol.~28, no.~3, p.~24, 2009.

\bibitem{irani1991improving}
M.~Irani and S.~Peleg, ``Improving resolution by image registration,''
  \emph{CVGIP: Graphical models and image processing}, vol.~53, no.~3, pp.
  231--239, 1991.

\end{thebibliography}
}
\end{document}